\documentclass[pdflatex,sn-nature]{sn-jnl}

\usepackage[utf8]{inputenc}
\usepackage{graphicx}
\usepackage{amsmath,amssymb}
\usepackage{booktabs}
\usepackage{array}  %
\usepackage{subcaption}  %
\usepackage{listings}
\usepackage[section]{placeins}
\providecommand{\orcidlink}[1]{\href{https://orcid.org/#1}{\textsuperscript{\textcolor[rgb]{0.651,0.808,0.224}{\textsf{\scriptsize iD}}}}\,}
\hypersetup{bookmarksdepth=subsection}

\graphicspath{{figures/}}

\begin{document}

\title[SoccerTrack v2]{SoccerTrack v2: A Full-Pitch Panoramic Video Dataset for Game State Reconstruction and Ball Action Spotting}

\author*[1]{\fnm{Atom} \sur{Scott}\,\orcidlink{0000-0003-2500-1096}}\email{atom.james.scott@gmail.com}

\author[2]{\fnm{Ikuma} \sur{Uchida}\,\orcidlink{0009-0000-2645-4455}}

\author[2]{\fnm{Kento} \sur{Kuroda}\,\orcidlink{0009-0006-0336-9673}}

\author[2]{\fnm{Yufi} \sur{Kim}}

\author[1]{\fnm{Keisuke} \sur{Fujii}\,\orcidlink{0000-0001-5487-4297}}

\affil*[1]{\orgname{Nagoya University}, \orgaddress{\country{Japan}}}

\affil[2]{\orgname{University of Tsukuba}, \orgaddress{\country{Japan}}}

\abstract{Soccer analytics draws on two kinds of information: spatio-temporal
data describing where players and the ball are, and event data describing what
they do. Public datasets offer them apart, or together only on broadcast
footage that leaves players outside the frame unobserved. SoccerTrack
v2 combines continuous full-pitch video, long player trajectories and
actor-linked events in one resource: ten university-level matches, 932 minutes
of fixed-camera 4K panoramic video, annotated per frame with metric pitch
coordinates, jersey numbers and persistent identities, roles and team sides for all players,
and with ball action events in twelve classes, linked to the acting players
through the same identifiers used in the trajectories. We fix a match-level
split and report baselines for two tasks. For game state reconstruction,
we run a full pipeline over all twenty halves and find that GS-HOTA scores
degrade as sequence length increases. For ball action spotting, we train a
model on the player trajectories, with and without the ball track. The data,
the split and the evaluation tooling are released so that both tasks can be
developed and compared at match length on the same footage.}


\maketitle

\section{Introduction}
\label{sec:intro}

Most quantities in soccer analytics are defined over positions and actions
together. Off-ball movement is valued relative to where the ball is and where it can
next be played~\cite{fernandez2018wide,spearman2018beyond}, and the value
of a possession asks how a spatial configuration changed as a consequence of
a specific action~\cite{decroos2019actions,fernandez2021framework}. Positions come
from tracking data and actions from event data, and both are in principle
recoverable from video, for which the vision stack has matured: detectors and multi-object trackers assign consistent boxes to
players~\cite{rfdetr25,bytetrack22,botsort22}, pitch line and keypoint
estimators project those boxes into metric
coordinates~\cite{deeplabv317,gutierrez2024nobells,broadtrack}, appearance models
and jersey number readers attach durable
identities~\cite{clipreid22,prtreid23,somers2024keypoint,parseq22,atienza2021vitstr,koshkina2024general}, and video
backbones with temporal heads localise actions in
time~\cite{i3d17,mvit22,hong2022spotting,tdeed24,astra23}. Game state reconstruction unifies the
geometric half of that pipeline into one task and one
metric~\cite{soccergsr24,luiten2021hota}.

Video plays two roles in this programme. It is the input from which such
systems must recover positions and events, and it is the reference against
which their output, and the annotations used to train and score them, can be
inspected: a position or an event released together with the footage it
describes can be rewound and looked at, whereas the same record released
without footage supports statistical modelling but not inspection. How much of
the scene the footage shows, and for how long, therefore shapes what a dataset
can be used for.

Existing public datasets each supply some of these ingredients. The
largest are broadcast-based. SoccerNet releases full broadcast matches with
action timestamps~\cite{giancola2018soccernet,deliege2021soccernet,cioppa2024soccernet},
automatically estimated camera calibration and positions of the visible
persons~\cite{cioppa2021camera}, and image-space player tracking on
30-second clips, with one 45-minute half held out for its
challenge~\cite{cioppa2022soccernet}; SoccerNet-GSR
defines game state reconstruction on 30-second broadcast
clips~\cite{soccergsr24}; SportsMOT tracks players in broadcast clips of three
sports~\cite{cui2023sportsmot}; MultiSports localises the actions of
individual players in broadcast clips of four sports, including
football~\cite{li2021multisports}; SoccerNet-GAR pairs broadcast event clips with
all-player tracking and group-activity labels~\cite{karki2026gar}; and FOOTPASS
aligns 54 full broadcast matches with player-linked play-by-play events and
all-player game state~\cite{ochin2026footpass}. SoccerNet's ball action
spotting task has since added attribution to the acting
team~\cite{giancola2025soccernet} and, on FOOTPASS, to the acting
player~\cite{cioppa2026soccernet}. In all of these the camera
follows play, so players outside the frame are either unannotated or have
their state reconstructed rather than observed (FOOTPASS reports 9 to 12 of the
22 players visible on average and imputes the remainder), and the tracking
benchmarks among them (SoccerNet tracking, SoccerNet-GSR, SportsMOT and
SoccerNet-GAR) are defined on clips rather than matches. Tracking and
event data are also released without video: Bassek et al.~\cite{bassek2025}
synchronise the optically tracked positions of every player with hierarchically
structured events across seven matches of the first and second German
divisions but cannot release the footage for licensing reasons; data
providers publish samples of the same kind, broadcast-derived tracking with
events for all 64 matches of the 2022 World Cup~\cite{pff2024worldcup} and
for twenty A-League matches~\cite{skillcorner2025opendata}, and tracking with
events for three anonymised matches~\cite{metrica2020sampledata}; and large
public event collections carry event
locations but no player tracking or video~\cite{pappalardo2019public}.
Instrumented full-pitch captures pair imagery with a separate measurement: the
Alfheim dataset combines a stitched panorama with body-worn sensors for the
home team~\cite{pettersen2014soccer}, and WorldPose recovers 3D poses from a
static multi-view rig and releases broadcast clips~\cite{worldpose24}; neither
carries events. Earlier full-pitch video releases, including our own
SoccerTrack v1 and TeamTrack~\cite{scott2022soccertrack,scott2024teamtrack},
provide positions on short clips without events, jersey numbers or roles, and
synthetic generators render video, trajectories and events together for
simulated rather than recorded play~\cite{soccer20}.
Section~\ref{subsec:comparison} tabulates the soccer releases among these.

SoccerTrack v2 adds the combination: continuous full-pitch video, long player
trajectories and actor-linked events in one resource, at match length. It
comprises ten university-level matches, 932 minutes of 4K panoramic video
recorded by fixed camera systems whose field of view spans the entire field of
play and its surroundings, so that players are not lost to camera motion as
play moves. Each frame
carries game state labels for the players on the pitch: metric pitch coordinates
on a canonical $105 \times 68$\,m pitch,
jersey numbers and a persistent player identifier, role, and team side,
with the median track annotated for 45 continuous minutes. The same footage
carries ball action labels over twelve event classes, and each event names the
acting team and, where the actor is identifiable, the actor in the \emph{same}
identity space as the tracks, so that an event resolves to a specific tracked
player at a metric pitch location. A curated bounding-box subset supports
multi-object tracking and was released with the SoccerTrack Challenge
2025~\cite{stc2025}.

The combination supports questions its parts do not. Because identities are
annotated for whole halves, a reconstruction method can be evaluated on
whether it holds identity over 45 minutes rather than 30 seconds, and the
effect of sequence length on its accuracy can be measured on identical
footage. Because events and state are synchronised, how much of the event
layer is determined by the configuration of players and how much by the ball,
and how a reconstruction system and an event spotter compose from video to
actions, can be measured on the same matches; because events name their
actors in the trajectory identity space, actor attribution can be scored
against a tracked player. And because the field of view covers the pitch, a
predicted or annotated position can be inspected against the footage at the
instant it is claimed, which is what failure analysis and label checking
require.

We make two contributions:

\begin{itemize}
  \item \textbf{Dataset.} Ten matches, two 45-minute halves each, of
        full-pitch panoramic video with per-frame game state for all players
        and ball actions linked to the players who performed them, released
        with a fixed match-level split, loaders and evaluation tools: one
        resource carrying a game state reconstruction benchmark and a ball
        action spotting benchmark on the same full-pitch footage.
  \item \textbf{Baselines.} Reference systems for both tasks: a game state
        reconstruction pipeline for fixed panoramic footage, run end to end
        over all twenty 45-minute halves, with a sequence-length study that
        rescores every half over growing windows and reruns one half on
        nested prefixes; and a ball action spotting model trained on the supplied
        trajectories and evaluated by cross-match cross-validation. In every half the
        reconstruction baseline's full-length score lies below its 30-second
        score, evidence that clip-scale evaluation can overstate match-length
        accuracy.
\end{itemize}

\section{Results}
\label{sec:results}

\FloatBarrier
\subsection{The SoccerTrack v2 dataset}
\label{subsec:dataset_overview}

\begin{figure}[!htbp]
  \centering
  \begin{subfigure}[b]{\linewidth}
    \centering
    \includegraphics[width=\linewidth]{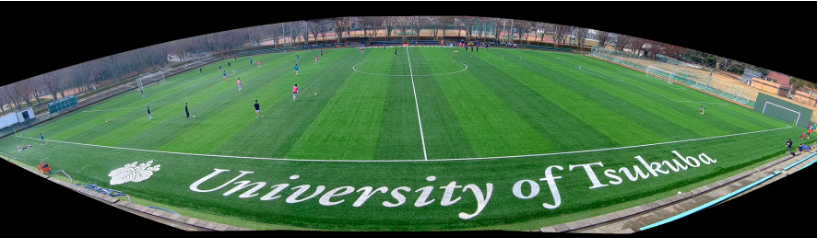}
    \caption{Daylight fixture.}
    \label{fig:1-1}
  \end{subfigure}
  \vspace{0.5em}
  \begin{subfigure}[b]{\linewidth}
    \centering
    \includegraphics[width=\linewidth]{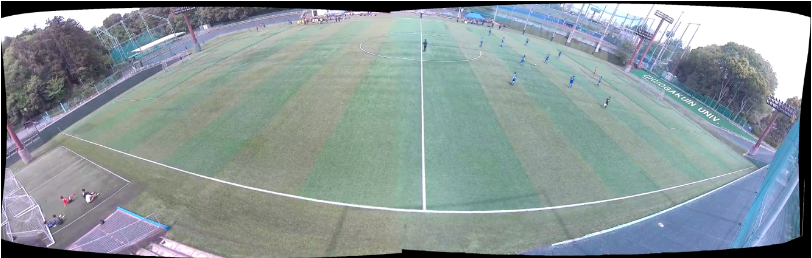}
    \caption{Daylight fixture at a second venue.}
    \label{fig:1-2}
  \end{subfigure}
  \vspace{0.5em}
  \begin{subfigure}[b]{\linewidth}
    \centering
    \includegraphics[width=\linewidth]{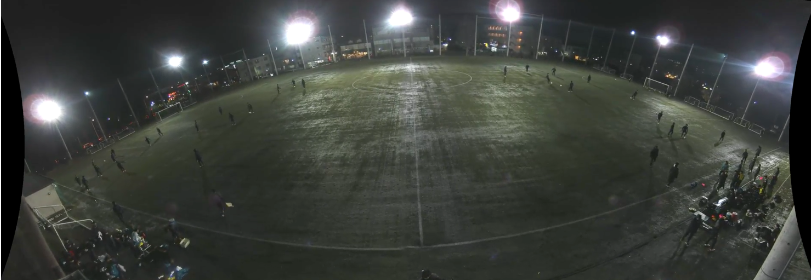}
    \caption{Night fixture under floodlights.}
    \label{fig:1-3}
  \end{subfigure}
  \caption{Example panoramic frames from SoccerTrack v2: the released field
  of view spans the pitch across venues and lighting conditions.}
  \label{fig:panoramic_views}
\end{figure}

SoccerTrack v2 comprises ten university-level matches, each released as two
45-minute periods, referred to below as halves, recorded by fixed
panoramic camera systems whose field of view spans the entire pitch. Play is
not lost to camera motion as it moves, as it is when a broadcast camera pans.
The corpus
totals $1.4$ million annotated frames, $932\,\text{min}$ of play at
$25\,\text{fps}$, and $30.7$ million annotated player-frame records, one
record per player annotated in a frame, about $22$ per frame. Recording
conditions vary across venues, weather and
lighting, including night fixtures (Fig.~\ref{fig:panoramic_views});
Table~\ref{tab:bas_dataset} gives the duration and event count of each
match.

\begin{figure}[!htbp]
  \centering
  \includegraphics[width=\linewidth]{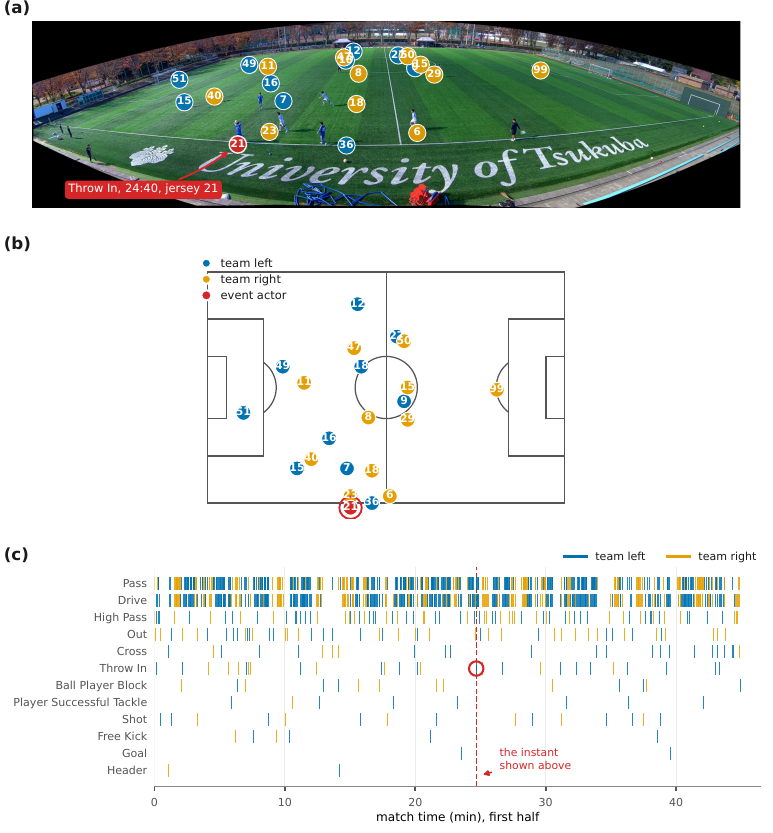}
  \caption{One match at a glance: M2, first half. (a) A frame of the
  released panoramic video at the instant of a \emph{Throw In} event, 24:40
  into the half: all twenty-two players are annotated, drawn as markers
  carrying their jersey numbers, with the acting player named by the event
  (jersey 21) in red. (b) The
  game state at the same frame, drawn from the game state annotations: all
  twenty-two players in metric pitch coordinates, coloured by team side, the
  event's actor at $(-10.5, 35.4)$\,m, beyond the touchline to take the
  throw. (c) The ball action layer of the same half: all $1{,}138$ events
  over 45 continuous minutes, one row per class, coloured by acting team,
  with the pictured instant marked. The events name their actors in the same
  identity space as the tracks, so the three views resolve to the same
  players at the same instants.}
  \label{fig:linkage}
\end{figure}

\begin{figure}[!htbp]
  \centering
  \includegraphics[width=\linewidth]{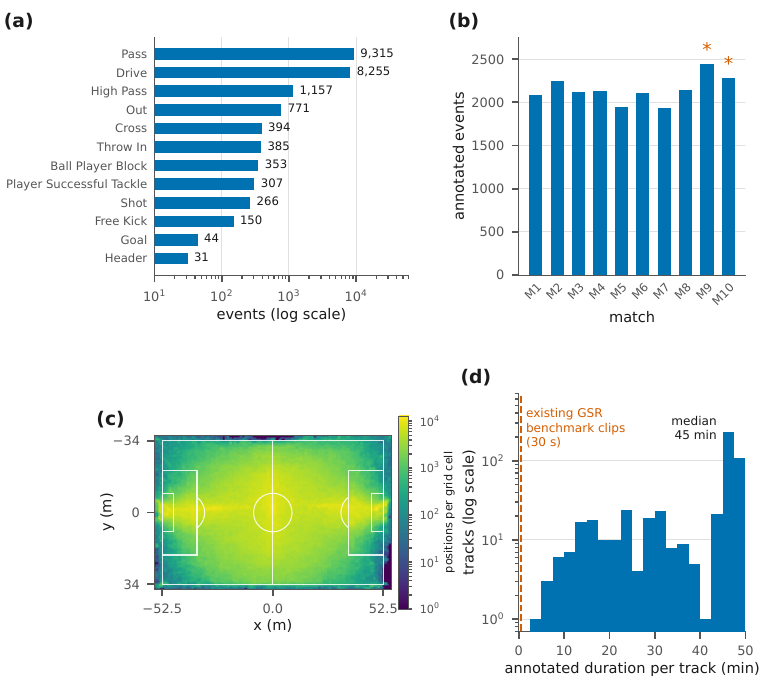}
  \caption{Dataset statistics. (a) Ball action class distribution over the
  $21{,}428$-event benchmark (log scale): the two most frequent classes account
  for $82\%$ of events and the support spans a factor of $300$. (b) Ball
  action events per match; asterisks mark the two matches whose recorded ball
  track is clamped to the pitch rectangle (Table~\ref{tab:bas_dataset}).
  (c) Density of all $30.7$ million annotated player positions of the game
  state layer in metric pitch coordinates, pooled over the ten matches (log
  colour scale). Positions are annotated on a $100 \times 100$ grid of the
  pitch, so each histogram cell measures $1.05 \times 0.68$\,m; $0.2\%$ of
  positions fall outside the displayed margin. (d) How long identities stay
  annotated. Each of the $522$ ground-truth tracks follows one player through
  one half; the histogram counts tracks by their annotated duration (log
  count scale). The median track is annotated for $45$ continuous minutes,
  while the dashed line marks the $30$-second clips on which existing game
  state benchmarks are defined~\cite{soccergsr24}.}
  \label{fig:stats}
\end{figure}

Each frame carries game state labels: metric pitch coordinates, jersey numbers
and persistent player identities, roles, and team sides. The same
footage carries ball actions over twelve classes, $21{,}428$ events, and
$99.5\%$ of them name the acting player in the identity
space of the tracks, so that an event resolves to a tracked player at a pitch
location (Fig.~\ref{fig:linkage}). This synchronisation of the two layers is
what the trajectory-based spotter of Section~\ref{subsec:bas_results}
consumes. Fig.~\ref{fig:stats}
summarises both layers: the class distribution spans a factor of $300$, the
positions cover the full pitch, and identities are held for whole halves, the
median track persisting for $45$ minutes where existing game state benchmarks
are defined on $30$-second clips~\cite{soccergsr24}.

\begin{table}[!htbp]
  \centering
  \caption{The ten SoccerTrack v2 matches, labelled M1 to M10 throughout
  the paper. \emph{Duration} is the annotated play time of each match at
  $25\,\text{fps}$. Every frame carries game state annotations for all
  players on the pitch: metric pitch coordinates on a canonical
  $105 \times 68\,\text{m}$ system, jersey number and persistent player
  identity, role, and team side. \emph{Events} counts the twelve-class ball
  actions. A curated tracking subset, one
  four-minute clip per match with every player boxed in every frame
  ($59{,}842$ frames and $1.3$ million boxes in total), is described in
  Section~\ref{subsec:methods_mot}. $\ast$ marks the two matches whose
  recorded ball track is clamped to the pitch rectangle and therefore never
  leaves play.}
  \label{tab:bas_dataset}
  \small
  \begin{tabular*}{\textwidth}{@{\extracolsep{\fill}}lrr@{}}
    \toprule
    Match & Duration (min) & Events \\
    \midrule
    M1 & 90 & 2,085 \\
    M2 & 92 & 2,252 \\
    M3 & 97 & 2,117 \\
    M4 & 97 & 2,129 \\
    M5 & 97 & 1,945 \\
    M6 & 94 & 2,106 \\
    M7 & 92 & 1,932 \\
    M8 & 91 & 2,144 \\
    M9$^\ast$ & 91 & 2,440 \\
    M10$^\ast$ & 91 & 2,278 \\
    \midrule
    Total & 932 & 21,428 \\
    \bottomrule
  \end{tabular*}
\end{table}

\FloatBarrier
\subsection{Positioning against existing datasets}
\label{subsec:comparison}

Table~\ref{tab:dataset-comparison} places the release beside the datasets of
Section~\ref{sec:intro}, grouped by the relation between the released video
and the game state. The broadcast-based group releases video and, in FOOTPASS,
actor-linked events with all-player state, but the state of players outside
the frame is imputed rather than observed: FOOTPASS reports 9 to 12 of 22
players visible on average~\cite{ochin2026footpass}, SoccerNet-GSR annotates
the visible persons of 30-second clips~\cite{soccergsr24}, and SoccerNet-GAR
refines all-player tracking from 4.5-second event clips and labels group
activities rather than actors~\cite{karki2026gar}. The instrumented group
observes every player with a measurement separate from the released camera:
Bassek et al.~\cite{bassek2025} withhold the video, the Alfheim
dataset~\cite{pettersen2014soccer} releases a full-pitch panorama with sensor
positions for the home team only and no events, and
WorldPose~\cite{worldpose24} releases poses and broadcast clips without
events. Providers' tracking and event
samples~\cite{pff2024worldcup,skillcorner2025opendata,metrica2020sampledata}
and event feeds~\cite{pappalardo2019public} release no video, and the
broadcast-derived samples estimate the players off screen, as FOOTPASS does;
SoccER~\cite{soccer20} renders the scene it annotates. Among full-pitch video
releases, SoccerTrack v1~\cite{scott2022soccertrack} and
TeamTrack~\cite{scott2024teamtrack} provide positions on clips without
events. SoccerTrack v2 is the entry that carries match-length full-pitch
video, all-player state within the released field of view and actor-linked
events together.

\begin{table}[!htbp]
\caption{Positioning of SoccerTrack v2. The first four groups are defined by
where the game state comes from relative to the released video: inferred
from the released broadcast, measured by an instrument separate from the
released camera, released without video, or rendered. The final group is
defined by footage instead: releases of full-pitch video, with this release
last; whether its provider measured the positions on the released view is
not documented (Section~\ref{subsec:methods_recording}). \emph{Events} counts released event
annotations and states whether each names the acting player. \emph{Length}
is as each release reports it.}
\label{tab:dataset-comparison}
\footnotesize
\begin{tabular}{@{}>{\raggedright\arraybackslash}p{0.17\textwidth}>{\raggedright\arraybackslash}p{0.19\textwidth}>{\raggedright\arraybackslash}p{0.19\textwidth}>{\raggedright\arraybackslash}p{0.15\textwidth}>{\raggedright\arraybackslash}p{0.16\textwidth}@{}}
\toprule
Dataset & Video released & Pitch state & Events & Length \\
\midrule
SoccerNet
  \cite{giancola2018soccernet,deliege2021soccernet,cioppa2022soccernet,cioppa2021camera}
  & broadcast, full matches & estimated positions of visible persons, no
  identities; image-space boxes on 30\,s clips
  & 110,458, not actor-linked
  & 500 matches, 764\,h \\
SoccerNet-GSR \cite{soccergsr24} & broadcast, 30\,s clips
  & visible persons only & none & 200 clips, 100\,min \\
SportsMOT \cite{cui2023sportsmot} & broadcast clips, three sports & no
  & none & 240 clips, 150\,k frames \\
SoccerNet-GAR \cite{karki2026gar} & broadcast, 4.5\,s event clips
  & all 22 and ball, refined from broadcast & 87,939, group-level
  & 64 matches, event windows only \\
FOOTPASS \cite{ochin2026footpass} & broadcast, full matches (NDA)
  & all 22, partly imputed (9 to 12 visible) & 102,992, actor-linked
  & 54 matches \\
\midrule
Bassek et al. \cite{bassek2025} & withheld & all 22, observed at 25\,Hz
  & 11,137, actor-linked & 7 matches \\
Alfheim \cite{pettersen2014soccer} & full-pitch panorama
  & home team only, sensors at 20\,Hz & none & 3 matches, ${\sim}220$\,min \\
WorldPose \cite{worldpose24} & broadcast clips & 3D poses, visible players
  & none & 88 clips, 49.7\,min \\
\midrule
Provider samples \cite{pff2024worldcup,skillcorner2025opendata,metrica2020sampledata}
  & no & all 22 and ball; broadcast-derived with off-screen players
  estimated in \cite{pff2024worldcup,skillcorner2025opendata}
  & actor-linked, counts not reported & 64, 20 and 3 matches \\
Event feeds \cite{pappalardo2019public} & no & event locations only
  & 3.25\,M, actor-linked & 1,941 matches \\
\midrule
SoccER \cite{soccer20} & rendered, full matches & all 22, simulated
  & simulated, actor-linked & 8 matches, 500\,min \\
\midrule
SoccerTrack v1 \cite{scott2022soccertrack}
  & full-pitch fish-eye and drone, 30\,s clips & all 22, GNSS at 10\,Hz
  & none & 63\,min of clips \\
TeamTrack \cite{scott2024teamtrack}
  & full-pitch fish-eye and drone, three sports & projected from boxes
  & none & 156\,min, 280\,k frames \\
\textbf{SoccerTrack v2 (ours)} & \textbf{full-pitch panorama, full matches}
  & \textbf{all players on the pitch, within the released field of view} & \textbf{21,428, actor-linked (99.5\%)}
  & \textbf{10 matches, 932\,min, 1.4\,M frames} \\
\bottomrule
\end{tabular}
\end{table}

\FloatBarrier
\subsection{Game state reconstruction baseline}
\label{subsec:gsr_results}

The reconstruction baseline is a game state reconstruction (GSR) pipeline
assembled from published components for this footage
(Section~\ref{subsec:methods_gsr}): its calibration stage is fitted per match
from the hand-annotated pitch keypoints supplied with the dataset
(Fig.~\ref{fig:calibration}), the jersey-region
gate was relaxed and a detection-level team-assignment variant added, both
set by us on released footage, and implementation changes let it run at
match length (Section~\ref{subsec:methods_scale}). The scores below
describe an adapted reference
system on this footage; because the adaptation drew on released data,
including prefixes of a test half, they are not a held-out transfer
measurement.

For each of the twenty halves we run the pipeline independently, under the
same configuration, on the whole half as annotated ($45$ to $49$ minutes)
and on its opening 30 seconds, the clip length of the SoccerNet-GSR
benchmark (Table~\ref{tab:gsr_results}). A full-half run costs about $37$
GPU-hours (Section~\ref{subsec:methods_scale}). Each result is scored both with the official
attribute-matching requirements and with attribute matching disabled; the
latter measures geometry and association alone.

\begin{table}[!htbp]
  \centering
  \caption{Game state reconstruction on all twenty halves, each run and
  scored over the whole half and, independently, on its opening 30
  seconds, the clip length of the SoccerNet-GSR benchmark: the game
  state reconstruction pipeline of Section~\ref{subsec:methods_gsr},
  configured by the authors on released footage that includes a
  test-split half and run end to end under one configuration, the
  per-detection team assignment. GS-HOTA is the official metric, which
  classes detections by (role, team, jersey); DetA, AssA and LocA are
  its detection, association and localisation components; \emph{Attrs
  off} rescores the same predictions with attribute matching disabled,
  so that only geometry and association count (components in
  Supplementary Table~\ref{tab:gsr_attrs_off}). The test split of the
  released match-level split is marked $\dagger$. Means are unweighted
  means of the per-half scores in each column, computed from unrounded
  scores.}
  \label{tab:gsr_results}
  \footnotesize
  \setlength{\tabcolsep}{4pt}
  \begin{tabular*}{\textwidth}{@{\extracolsep{\fill}}lrrrrrrr@{}}
    \toprule
    & \multicolumn{2}{c}{Opening 30\,s} & \multicolumn{5}{c}{Full half (45\,min)} \\
    \cmidrule(lr){2-3} \cmidrule(l){4-8}
    Half & GS-HOTA & Attrs off & GS-HOTA & DetA & AssA & LocA & Attrs off \\
    \midrule
    M1, 1st & 23.10 & 27.37 & 6.68 & 4.29 & 10.50 & 71.77 & 8.46 \\
    M1, 2nd & 6.49 & 31.31 & 3.60 & 1.15 & 11.25 & 72.06 & 7.44 \\
    M2, 1st & 55.20 & 66.12 & 14.28 & 10.04 & 20.37 & 83.02 & 22.14 \\
    M2, 2nd & 26.49 & 49.15 & 12.13 & 8.01 & 18.38 & 83.00 & 22.08 \\
    M3, 1st & 36.49 & 61.75 & 16.59 & 10.78 & 25.73 & 82.73 & 26.09 \\
    M3, 2nd & 32.49 & 63.71 & 12.21 & 6.83 & 21.84 & 85.71 & 23.71 \\
    M4, 1st & 24.60 & 48.57 & 12.36 & 7.11 & 21.58 & 86.42 & 24.56 \\
    M4, 2nd & 41.14 & 58.80 & 11.03 & 8.39 & 14.55 & 82.55 & 20.05 \\
    M5, 1st & 39.89 & 72.09 & 18.23 & 15.68 & 21.20 & 86.67 & 25.10 \\
    M5, 2nd & 39.74 & 62.52 & 16.54 & 12.26 & 22.34 & 87.00 & 24.68 \\
    M6, 1st & 29.59 & 54.83 & 13.36 & 10.47 & 17.08 & 85.33 & 23.00 \\
    M6, 2nd & 31.60 & 54.20 & 10.22 & 8.16 & 12.80 & 84.88 & 18.49 \\
    M7, 1st$^\dagger$ & 37.15 & 77.45 & 18.09 & 14.67 & 22.32 & 86.84 & 26.04 \\
    M7, 2nd$^\dagger$ & 33.51 & 66.37 & 12.89 & 7.75 & 21.48 & 86.67 & 25.08 \\
    M8, 1st & 41.80 & 72.72 & 12.73 & 9.60 & 16.91 & 85.74 & 25.89 \\
    M8, 2nd & 34.29 & 60.62 & 12.26 & 7.80 & 19.34 & 83.17 & 21.65 \\
    M9, 1st$^\dagger$ & 21.56 & 29.53 & 9.92 & 9.69 & 10.16 & 79.04 & 16.15 \\
    M9, 2nd$^\dagger$ & 8.92 & 52.58 & 2.94 & 1.23 & 7.10 & 70.19 & 17.91 \\
    M10, 1st & 38.43 & 60.79 & 15.27 & 8.71 & 26.79 & 85.32 & 29.26 \\
    M10, 2nd & 24.41 & 43.92 & 13.64 & 8.07 & 23.07 & 86.18 & 25.40 \\
    \midrule
    Mean, all twenty halves & 31.35 & 55.72 & 12.25 & 8.53 & 18.24 & 82.71 & 21.66 \\
    Mean, test split$^\dagger$ & 25.29 & 56.48 & 10.96 & 8.34 & 15.26 & 80.68 & 21.29 \\
    \bottomrule
  \end{tabular*}
\end{table}

Over the twenty halves the full-half GS-HOTA is $12.25$ (DetA $8.53$,
AssA $18.24$, LocA $82.71$), and $21.66$ with attributes off; the four
halves of the released test split average $10.96$. On the opening 30
seconds the same twenty halves average $31.35$, and $55.72$ with attributes
off (Table~\ref{tab:gsr_results}). In every half the full-half score lies
below the 30-second score of the same half, by a factor of $1.8$ to $3.9$
under the official metric and $1.7$ to $4.2$ with attributes off
(Table~\ref{tab:gsr_results}). The 30-second windows are anchored at kickoff rather than sampled throughout
each half. Both lengths in Table~\ref{tab:gsr_results} use the
per-detection team-assignment variant; the comparison of team-assignment
variants below concerns only M7, first half. The sequence-length study
below examines the gap by rescoring every half over growing windows and by
rerunning the pipeline on one half.

The spread across halves is itself a finding. Over whole halves the official
score ranges from $18.23$ (M5, first half) to $2.94$ (M9, second
half), and the twenty halves divide by match: the four halves of M1 and
M9 score between $2.94$ and $9.92$, the sixteen halves of the other eight
matches between $10.22$ and $18.23$, and localisation under
the official metric separates the two groups without overlap ($70.2$ to
$79.0$ against $82.6$ to $87.0$). With attributes off, detection accuracy
(DetA, averaged over the metric's localisation thresholds, which span
$0.65$ to $5$\,m) lies between $15.9$ and $41.4$ on the four halves against
$42.5$ to $51.2$ on the sixteen (Supplementary Table~\ref{tab:gsr_attrs_off}
lists the attributes-off components). On two of the four, the second halves
of M1 and M9, attribute matching imposes a substantial additional
penalty (official DetA $1.15$ and $1.23$), taking the official score to
$3.60$ and $2.94$ where geometry and association alone score $7.44$ and
$17.91$. On M7, the attributes-off score changes less between halves
($26.04$ against $25.08$) than the official score ($18.09$ against $12.89$).
A related separation is visible in the first 30 seconds: three of the four
halves of M1 and M9 score between $27.37$ and $31.31$ with attributes off, and their
localisation is about ten points lower on average ($73$ to $78$ against $79$
to $90$, Supplementary Table~\ref{tab:gsr_attrs_off}); the other seventeen,
among them M9, second half ($52.58$), score between $43.92$ and $77.45$.
The four lowest full-half scores belong to two matches, and several factors
vary together across them: capture conditions, calibration quality (the
misaligned stitch seam of M1, Section~\ref{subsec:methods_calibration}),
attribute legibility, and the ground-truth property noted in Methods
(player and ball positions clamped to the pitch rectangle in M9). M10, whose positions are
clamped in the same way, scores within the stronger group, so clamping alone
does not account for the divide. We do not
separate these potential contributors here, which is one reason we report
every half rather than a single aggregate.

\paragraph{With attributes off, association is the larger bottleneck in every half.}
The association component is the smaller of the two in
every half: AssA lies between $3.3$ and $18.6$ against DetA between $15.9$
and $51.2$, a ratio of $2.5$ to $6.0$ (Supplementary
Table~\ref{tab:gsr_attrs_off}), and the pipeline produces $123$ to $422$
tracklets per half against $22$ to $35$ annotated identities, where the
same halves at 30 seconds produce $23$ to $37$ tracklets for $22$.
Table~\ref{tab:gsr_attributes} decomposes the score of
M7 first half, the best half of the test split, in detail. Disabling
attribute matching gives $26.04$, the ceiling for attribute correction with the
current detections and associations held fixed: perfect role, team and
jersey would score exactly that, so all attribute work together is worth at
most $7.94$ points. Within that budget the successive costs, in the
displayed order, are role $-0.40$, team $-3.27$ and jersey $-4.27$, and the per-attribute accuracies on $36{,}665$
geometrically matched pairs are role $97.92\%$, team $76.17\%$ and jersey
$45.95\%$. Jersey numbers are wrong rather than missing: only $3.4\%$ of
predictions carry a null jersey.

The ceiling itself is set by detection in pitch space and by association,
and these are far from balanced. With attributes off the pipeline places the
players it detects accurately ($\text{LocA} = 84.7$), reaches a detection
accuracy of $\text{DetA} = 51.2$ and fails to
hold identity over time ($\text{AssA} = 13.4$): perfect detection
would raise the ceiling to $36.6$, perfect association to $71.6$. DetA here
confounds missed and spurious detections with projection error between the
provider's calibration and the keypoint homography, which are consistent
only to $2$ to $3$\,m (Section~\ref{subsec:methods_calibration}); we have
not separated the two. Nor is the association loss simple fragmentation. The
run produces $345$ predicted tracklets for $23$ identities, but most are
short fragments: discarding every tracklet shorter than $100$ detections
leaves $65$ and changes GS-HOTA by $0.01$, while $41$ of the $154$ tracklets
with at least ten detections span more than three times as many frames as
they contain detections, so identities are reused across gaps rather than
following one player. On this half, then, the association loss is identity reuse across gaps
more than fragmentation.

\begin{table}[!htbp]
  \centering
  \caption{Marginal cost of each game state attribute, computed on the first
  half of match M7 (the best-performing half of the test split in
  Table~\ref{tab:gsr_results}). GS-HOTA classes detections by the triple
  (role, team, jersey), so a detection with any incorrect attribute cannot
  match ground truth at all; each row switches one further attribute on, and
  \emph{Marginal} is the incremental cost of enabling that attribute after
  those in the preceding row, computed from unrounded scores, so it may
  differ by $0.01$ from the difference of the
  displayed values. DetA and AssA are the detection and association
  components of the metric. The first row, with no attributes scored, is the
  ceiling available to the official metric: perfect attributes would score
  exactly this, so it carries no marginal cost.}
  \label{tab:gsr_attributes}
  \small
  \begin{tabular*}{\textwidth}{@{\extracolsep{\fill}}lrrrr@{}}
    \toprule
    Configuration & GS-HOTA & DetA & AssA & Marginal \\
    \midrule
    Geometry and association only & \textbf{26.04} & 51.21 & 13.41 & \multicolumn{1}{c}{ceiling} \\
    {+} role & 25.64 & 49.36 & 13.48 & $-0.40$ \\
    {+} role, team & 22.36 & 35.12 & 14.37 & $-3.27$ \\
    {+} role, team, jersey (official) & \textbf{18.09} & 14.67 & 22.32 & $-4.27$ \\
    \bottomrule
  \end{tabular*}
\end{table}

\begin{figure}[!htbp]
  \centering
  \includegraphics[width=\linewidth]{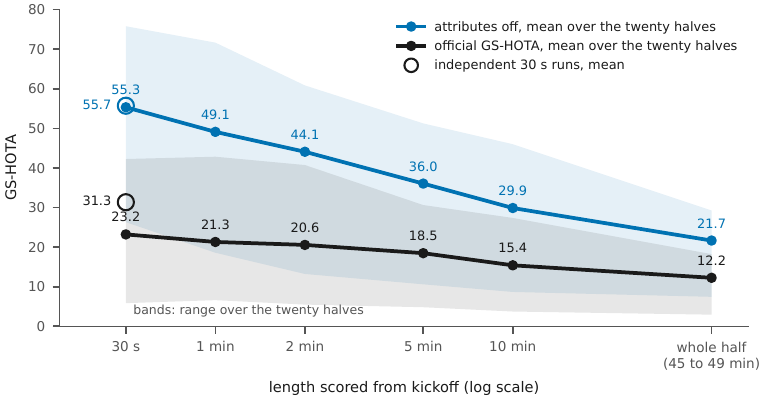}
  \caption{Game state reconstruction accuracy against the length scored.
  Filled markers: the whole-half predictions of every half
  (Table~\ref{tab:gsr_results}) scored over their first 30 seconds, 1, 2,
  5 and 10 minutes and the whole half, mean over the twenty halves
  (Supplementary Table~\ref{tab:gsr_windows}), under the official metric
  (black) and with attribute matching disabled (blue); the bands span the
  twenty halves. Hollow markers: the mean of the independent 30-second
  runs of Table~\ref{tab:gsr_results}, in which the pipeline saw only the
  clip.}
  \label{fig:gsr_length}
\end{figure}

\paragraph{Clip-scale scores exceed full-half scores on this footage.}
Two experiments separate how far this gap belongs to the length scored and
how far to running the pipeline on less footage. In the first, the
whole-half predictions of every half are held fixed and scored over growing
windows from kickoff, from the first 30 seconds to the whole half
(Fig.~\ref{fig:gsr_length}; Supplementary Table~\ref{tab:gsr_windows}
lists every half and window). The mean over the twenty
halves falls monotonically, from $23.21$ at 30 seconds to $12.25$ over the
whole half under the official metric and from $55.33$ to $21.66$ with
attributes off. Individual halves wobble, and on two, M6 first half and M10
second half, the whole half scores above its 30-second window ($13.36$
against $11.90$ and $13.64$ against $11.09$); on every other half the whole
half scores below it. The independent 30-second
runs of Table~\ref{tab:gsr_results} score higher still on the same clips,
$31.35$ against $23.21$ under the official metric, while with attributes off
the two agree ($55.72$ against $55.33$): running the pipeline on a 30-second
clip rather than on the whole half gains about eight points, almost
entirely in the attributes, which the pipeline assigns per tracklet and by
clustering over the whole sequence (Section~\ref{subsec:methods_gsr}).

In the second experiment the pipeline is rerun end to end on nested prefixes
of the first half of match M7, the half on which the adaptation was also
explored, under the per-detection team-assignment configuration, a design
the fixed cameras and continuous footage permit, so that footage,
calibration and configuration are shared across prefixes and each prefix is
a separate run. Accuracy peaks at one minute, at a GS-HOTA of $47.59$, then
falls monotonically through every longer prefix ($39.88$ at two, $30.97$ at
five and $25.60$ at ten minutes), reaching $18.09$ over the full
half: $38\%$ of the one-minute figure from the same half. From the
30-second prefix to the full half, predicted tracklets grow from 26 to 345 against 22 real identities in
the shorter prefixes and 23 over the full half, a substitute entering late.

Two properties of the design bear on both experiments: the windows are anchored
at kickoff, and the metric's association term is computed over whole
trajectories, so some decline with length is intrinsic to scoring longer
horizons. The first experiment measures that intrinsic decline directly,
since its predictions do not change; the second adds the effect of running
on less footage. The increase from 26 to 345 predicted tracklets is
consistent with fragmentation but does not isolate its contribution from
either. A method characterised on thirty-second or one-minute excerpts, the
lengths at which existing game state benchmarks are
defined~\cite{soccergsr24,cui2023sportsmot}, cannot therefore be assumed to
retain that accuracy over a half: full-match evaluation is a distinct
protocol, not a longer clip.

\paragraph{The best configuration depends on the length evaluated.}
Team assignment makes the same point from a different direction, on the same
prefixes. Clustering per-detection appearance embeddings, rather than
tracklet-mean embeddings, raises team accuracy at five
minutes from $51.6\%$, chance on two classes, to $86.6\%$, and GS-HOTA from
$19.48$ to $30.97$. At two minutes and below the ranking reverses:
tracklet-mean clustering scores $40.35$ against $37.15$ at thirty seconds,
$57.82$ against $47.59$ at one minute and $44.81$ against $39.88$ at two
minutes. No single variant is best at every length, so Methods describes
both; every score of Table~\ref{tab:gsr_results}
uses the per-detection variant, and the crossover is a concrete instance of
a design choice that looks right on clips but no longer holds by five
minutes, the longest prefix at which both variants were scored.

\FloatBarrier
\subsection{Ball action spotting baseline}
\label{subsec:bas_results}

The second baseline spots ball actions (ball action spotting, BAS) from the
released game state rather than from pixels, which the synchronised event and
state layers permit (Section~\ref{subsec:dataset_overview}). It is an offline
temporal convolutional network over per-frame features of the supplied player
trajectories, with or without the supplied ball track
(Section~\ref{subsec:methods_bas_traj}, Fig.~\ref{fig:baselines}b).
Events have been detected from tracking data with rule-based algorithms,
classical classifiers and learned
models~\cite{richly2016recognizing,link2017individual,khaustov2020recognizing,fassmeyer2021toward,vidalcodina2022automatic},
and group activities from trajectories and, separately, from video, with most
events detectable from either~\cite{sanford2020group}; SoccerNet-GAR compares
the two inputs under one protocol and finds tracking the stronger for group
activity recognition~\cite{karki2026gar}. Here the trajectories come with full-pitch
video of the same matches and are scored under the ball action spotting
metric. Because the input is the supplied
ground-truth state, the figures measure how much of the event layer the
released state carries under one offline model; spotting from video works
from a different input, and its obstacles on this footage are set out in
Section~\ref{sec:discussion}. The supplied ball track is itself a
reconstruction rather than a per-frame measurement: the ball moves in
straight lines at constant speed between anchors, most of which are
annotated events, so most of its changes of velocity fall within about
$0.1$\,s of an annotated event instant. The with-ball
condition therefore sees information tied to the event annotation, which may
inflate its scores relative to an observed ball, and we report it as a
reference condition rather than as the value of ball tracking.

Table~\ref{tab:bas_cv} reports the five-fold cross-match evaluation and
Table~\ref{tab:bas_per_class} the per-class breakdown pooled over all ten
matches. Player positions alone yield a cross-validated mean average
precision of $0.222 \pm 0.021$ at $\tau = 1$\,s and $0.524 \pm 0.048$ at
$\tau = 5$\,s, against $0.008$ and $0.064$ for a uniform-cadence chance
baseline. Adding the supplied ball track raises these to $0.599 \pm 0.064$
and $0.666 \pm 0.045$; on the Challenge fold alone, the held-out pair of the
match-level split, the with-ball figure is $0.617$. The chance baseline is
not a formality: its support-weighted score at $\tau = 5$\,s is $0.279$,
because two classes account for $82\%$ of events and, over the $932$
minutes of Table~\ref{tab:bas_dataset}, an event occurs every $2.6$\,s and a
pass every $6$\,s, so the margin over chance rather than the absolute value
is the quantity of interest.

The two conditions separate far more at the tight tolerance than at the loose
one. Without the ball, tightening $\tau$ from $5$\,s to $1$\,s costs $0.302$
mean average precision; with the ball it costs $0.067$. At five-second
tolerance the ball adds $0.14$ mean average precision ($0.524$ to $0.666$);
at one second it adds $0.38$, a factor of $2.7$. A larger improvement at
the tighter tolerance is what a track whose changes of velocity coincide
with the annotated events would produce, so it does not measure the temporal
precision an observed ball would provide.

\paragraph{Duel outcomes are the weakest classes in both conditions.}
Events defined by the outcome of a duel are the two weakest classes with and
without the ball: \texttt{Ball Player Block} reaches $0.070$ from
trajectories and $0.181$ with the ball, and \texttt{Player Successful
Tackle} $0.046$ and $0.121$, against $0.286$ and $0.853$ for \texttt{Pass}.
Both classes ask which of two converging players won a contact, and the
pattern is consistent with that outcome not being represented in ground
positions and ball location at this resolution. The classes the ball
transforms are those defined by the ball rather than by the players:
\texttt{Throw In} gains $0.66$, \texttt{Drive} $0.60$, \texttt{Pass} $0.57$
and \texttt{Free Kick} $0.53$. \texttt{Out} gains only $0.22$, reaching
$0.364$: it is a synthesised restart marker
(Section~\ref{subsec:methods_bas_annotation}) rather than an observed
boundary crossing, and in two of the ten matches the recorded ball never
crosses the lines. \texttt{Header} ($n = 31$) and
\texttt{Goal} ($n = 44$) are reported for completeness; at that support the
figures are indicative only, which is why both an unweighted and a
support-weighted mean are given.

\paragraph{The with-ball figures pool two regimes.}
The ball track also differs between matches. In
M7 the ball leaves the field of play on $3\%$ of frames, as it must; in
M9 and M10 the recorded position is clamped to the pitch rectangle
and never does. On the Challenge fold, adding the ball raises mean average
precision at $\tau = 1$\,s from $0.267$ to $0.734$ on M7 and from
$0.191$ to $0.530$ on M9. The pooled with-ball figure on that fold
($0.617$) therefore averages an unclamped and a clamped match;
Table~\ref{tab:bas_cv} gives both.

\begin{table}[!htbp]
  \centering
  \caption{Five-fold cross-match evaluation of the spotting model on the
  supplied trajectories. Each match appears in the test set exactly once;
  fold 0 is the SoccerTrack Challenge 2025 pair. Validation for each fold is
  the following fold's test pair, so no fold's decoding or stopping epoch is
  chosen on its own test matches; the network settings shared by all folds
  were fixed once on M2 and M10 (Section~\ref{subsec:methods_bas_traj}).
  Each entry is a single training run.
  The last block gives the two matches of the Challenge fold separately: the
  ball track is unclamped in M7 and clamped to the pitch rectangle in M9, so the
  value of the ball differs sharply between them and the fold's pooled row
  averages two regimes.}
  \label{tab:bas_cv}
  \small
  \begin{tabular*}{\textwidth}{@{\extracolsep{\fill}}llcccc@{}}
    \toprule
    & & \multicolumn{2}{c}{$\tau = 1\,$s} & \multicolumn{2}{c}{$\tau = 5\,$s} \\
    \cmidrule(lr){3-4}\cmidrule(lr){5-6}
    Input & Test fold & mAP & mAP$_w$ & mAP & mAP$_w$ \\
    \midrule
    trajectory & M7, M9\,(Challenge) & 0.207 & 0.202 & 0.472 & 0.637 \\
     & M1, M2 & 0.222 & 0.272 & 0.501 & 0.676 \\
     & M3, M4 & 0.231 & 0.259 & 0.600 & 0.709 \\
     & M5, M6 & 0.199 & 0.258 & 0.516 & 0.678 \\
     & M8, M10 & 0.252 & 0.275 & 0.531 & 0.664 \\
    \cmidrule(lr){2-6}
    & \textbf{Mean} & \textbf{0.222} & \textbf{0.253} & \textbf{0.524} & \textbf{0.673} \\
    & SD across folds & 0.021 & 0.030 & 0.048 & 0.026 \\
    \midrule
    trajectory + ball & M7, M9\,(Challenge) & 0.617 & 0.788 & 0.679 & 0.826 \\
     & M1, M2 & 0.633 & 0.827 & 0.685 & 0.846 \\
     & M3, M4 & 0.624 & 0.819 & 0.701 & 0.843 \\
     & M5, M6 & 0.636 & 0.808 & 0.679 & 0.832 \\
     & M8, M10 & 0.486 & 0.737 & 0.588 & 0.777 \\
    \cmidrule(lr){2-6}
    & \textbf{Mean} & \textbf{0.599} & \textbf{0.796} & \textbf{0.666} & \textbf{0.825} \\
    & SD across folds & 0.064 & 0.036 & 0.045 & 0.028 \\
    \midrule
    \multicolumn{6}{@{}l}{\emph{Challenge fold, per match}} \\
    trajectory & M7 & 0.267 & 0.275 & 0.550 & 0.669 \\
     & M9 & 0.191 & 0.176 & 0.437 & 0.632 \\
    trajectory + ball & M7 & 0.734 & 0.829 & 0.785 & 0.852 \\
     & M9 & 0.530 & 0.742 & 0.604 & 0.815 \\
    \midrule
    \textit{Uniform cadence} & Mean & 0.008 & 0.033 & 0.064 & 0.279 \\
    \bottomrule
  \end{tabular*}
\end{table}

\begin{table}[!htbp]
  \centering
  \caption{Per-class average precision at $\tau = 1\,$s, pooled over the five cross-match folds and therefore over all ten matches. $n$ is the number of ground-truth events across the benchmark. Rows marked $\dagger$ have $n < 100$. Classes are ordered by support. Because AP is pooled over matches, the column means differ slightly from the fold-averaged means of Table~\ref{tab:bas_cv}.}
  \label{tab:bas_per_class}
  \small
  \begin{tabular*}{\textwidth}{@{\extracolsep{\fill}}lrccc@{}}
    \toprule
    Class & $n$ & chance & trajectory & {+}ball \\
    \midrule
    Pass & 9315 & 0.042 & 0.286 & 0.853 \\
    Drive & 8255 & 0.036 & 0.227 & 0.830 \\
    High Pass & 1157 & 0.007 & 0.320 & 0.745 \\
    Out & 771 & 0.004 & 0.148 & 0.364 \\
    Cross & 394 & 0.001 & 0.327 & 0.794 \\
    Throw In & 385 & 0.006 & 0.254 & 0.910 \\
    Ball Player Block & 353 & 0.000 & 0.070 & 0.181 \\
    Player Successful Tackle & 307 & 0.002 & 0.046 & 0.121 \\
    Shot & 266 & 0.002 & 0.317 & 0.735 \\
    Free Kick & 150 & 0.001 & 0.171 & 0.701 \\
    Goal$^\dagger$ & 44 & 0.000 & 0.336 & 0.718 \\
    Header$^\dagger$ & 31 & 0.000 & 0.189 & 0.347 \\
    \bottomrule
  \end{tabular*}
\end{table}

\section{Discussion}
\label{sec:discussion}

SoccerTrack v2 places continuous full-pitch video, match-length trajectories
with persistent identities and actor-linked events in one resource. Its
value lies in that combination: methods that reason over positions and events
together can be trained and scored on the same matches, a reconstruction
system can be asked to hold identity for 45 minutes rather than 30 seconds,
and what either produces can be inspected against the footage at the moment
it is claimed to occur. The two baselines provide initial reference results
and identify limitations of the evaluated systems.

\paragraph{What the baselines reveal.}
Over whole halves, attributes-off detection accuracy (DetA) ranges from
$15.9$ to $51.2$ (Supplementary
Table~\ref{tab:gsr_attrs_off}), while official localisation exceeds $70$ in
every half. With attributes off, association is the smaller component in
every half (AssA $3.3$ to $18.6$ against DetA $15.9$ to $51.2$); on M7,
first half, perfect association would nearly triple the attributes-off
score, whereas perfect detection would add about eleven points. The four lowest full-half scores belong to two matches;
capture and calibration conditions, jersey legibility and ground-truth
properties vary together across them, and their contributions have not been
separated, although clamping alone does not explain the gap, since M10 is
clamped like M9 and scores with the stronger group. Three of the four weakest full halves also form the lowest-scoring
attributes-off group at 30 seconds. Scored over growing windows from
kickoff, the twenty-half mean of the whole-half runs falls from $23.21$ at
30 seconds to $12.25$ over the whole half, and independent 30-second runs
score higher still ($31.35$); on nested prefixes of one half rerun end to
end, accuracy peaks at one minute and falls to $38\%$ of that value over the
full half, and the better team-assignment variant changes with the length
evaluated. Together, these results
show that opening-clip scores overstate full-half scores for this adapted
baseline on this corpus, supporting full-half evaluation alongside
clip-scale evaluation. The spotting baseline shows the complementary
pattern on the supplied trajectories: at five-second tolerance player
positions alone recover most of what the with-ball model reaches ($0.524$
against $0.666$), while at a one-second tolerance adding the supplied ball
track increases mAP by a factor of $2.7$, a gain that the track's
reconstruction around the annotated events may inflate; both models
struggle with duel outcomes. These offline
results on ground-truth state quantify how well events can be predicted
from the supplied trajectories.

\paragraph{Limitations.}
The annotations originate from a commercial tracking and event system. We
inspected the positions against the video by projecting them onto the
panorama and onto a bird's-eye view of the pitch (Figs.~\ref{fig:linkage}
and~\ref{fig:calibration}), in spot checks and throughout the development of
the baselines, and the event timelines were checked and corrected against
the video during curation. We have not conducted a quantitative audit: there
is no inter-annotator agreement, no correction-rate statistic and no
positional residual against independently placed reference points. The
full-pitch view is what makes such inspection possible; it does not by itself
establish accuracy. What can be stated is that the release itself fixes a
scale of a few metres: positions are quantised to a $1.05 \times 0.68$\,m
grid, and the hand-annotated calibration correspondences are self-consistent
only to roughly $2$ to $3$\,m on the pitch
(Section~\ref{subsec:methods_calibration}); neither figure measures the
provider's accuracy, and positional claims finer than a few metres are not
supported. Player identity is carried by the jersey number and by a provider
identifier that persists across the halves of a match and, for squads that
appear in several fixtures, across matches
(Section~\ref{subsec:methods_gsr_annotation}); how the provider maintained
identity where a number was occluded is not documented, identity consistency
was not audited, and the pseudonymised release does not resolve identifiers
to persons. The field of view covers the pitch; it does not follow that a given
player is unoccluded at every instant, and occlusion by other players was
not audited. All ten matches are university-level fixtures
from a single league, recorded by one provider's fixed panoramic systems, so
playing style, tempo and camera conditions may not transfer to professional
or broadcast settings, and exhaustive bounding-box annotation was beyond the
resources of this release, so box-level supervision is released as a curated
subset (Section~\ref{subsec:methods_mot}).

The baselines are bounded in turn. Reconstruction is evaluated on all twenty
halves, over the whole half and on the opening 30 seconds; the attribute
decomposition and the nested-prefix reruns were computed on one half. The
adapted pipeline's jersey gate and team-assignment variant were chosen on
released footage that includes a test half, so its scores characterise the
adapted system on this data, and no variance estimate accompanies any
single-half figure. The spotting figures assume the supplied trajectories and
an offline spotter. The with-ball condition uses a ball track
reconstructed around the annotated events rather than measured per frame
(Section~\ref{subsec:methods_bas_traj}), and its cross-validated figures
include M9 and M10, whose recorded ball never leaves the pitch.

\paragraph{Relation to broadcast datasets.}
SoccerTrack v2 complements the broadcast-derived corpora that dominate the
field rather than replacing them. Models intended for television footage are
best trained and evaluated on television footage, and the SoccerNet
family~\cite{giancola2018soccernet,deliege2021soccernet,cioppa2022soccernet,soccergsr24}
remains the right resource for that setting. What a fixed panoramic recording
adds is a field of view that spans the pitch: players are not lost to camera
motion, every annotated position has footage at the instant it is claimed,
identities are held for whole halves (the
median track spans 45 minutes), and events resolve to those identities.
Whether the provider measured each position on this view is not documented
(Section~\ref{subsec:methods_recording}). The two settings answer different
questions, and the game state annotations use the SoccerNet GameState format
while the event annotations follow the SoccerNet ball action spotting
convention, so tooling and methods can move between them.

\paragraph{Spotting events from pixels.}
The spotting baseline consumes trajectories rather than video because of a
property of the footage. A panoramic frame covering the whole pitch at 4K
presented to a video backbone at its usual input resolution requires an
order-of-magnitude downscale, at which a 22\,cm ball spans about one pixel,
or else a crop. A crop centred on the annotated ball position is an oracle a
deployed system would not have; a crop centred on a detected ball must first
solve ball detection in panoramic footage, where the ball spans only a few
pixels, and compounds its input's failures at exactly the contested moments
where events concentrate. The released ball track cannot serve as ground truth
for such a detector: it is anchored at the annotated events and interpolated
between them rather than tracked, and it is clamped to the pitch rectangle
in two matches (Table~\ref{tab:bas_dataset}). A ball track in pitch
coordinates is in any case a demanding artefact, since an airborne ball
projected through a ground-plane homography lands far from where it is. The
remaining alternatives, a play-centred crop derived from
player positions or a full-frame downscale, are natural first experiments.

\paragraph{Outlook.}
The most informative next experiment is the composition of the two
baselines. The spotting figures assume the supplied game state, so running
the same spotter on tracks predicted by the reconstruction baseline would
show how much of the end-to-end gap is attributable to reconstruction and how
much to spotting, and would give the first measurement from panoramic video
to ball actions on this data. The fixed match-level split and the released
evaluation tooling are intended to make such results directly comparable
across contributions. The SoccerTrack Challenge leaderboard~\cite{stc2025},
rather than this paper, carries the reference tracking results on the
curated bounding-box subset. Because the panoramic
recording contains every narrower framing within it, the cost of
ball-following framing can also be studied on this data, by rendering a
viewport and comparing against the full view; we leave that ablation to
future work.

\section{Methods}
\label{sec:methods}

\subsection{Match recording and camera systems}
\label{subsec:methods_recording}

Matches were captured by fixed, stitched panoramic camera systems (BePro)
whose field of view covers the complete pitch; the released panorama is
about 4K wide ($3840$ or $4096$\,px) and $1080$ to $1906$\,px tall. The ten
matches involve university-level
amateur teams, selected for their standard of play and for their willingness to
participate in data collection. Positional data for players were provided by the
BePro system, synchronised to video timestamps; the provider's processing
chain, and whether it derives the positions from the same panoramic feed that
is released or from a separate pass over its cameras, is not part of the
release, so the positions are described here as lying within the released
field of view rather than as measured on the released view.

\subsection{Pitch coordinates and calibration}
\label{subsec:methods_calibration}

\begin{figure}[!htbp]
  \centering
  \includegraphics[width=\linewidth]{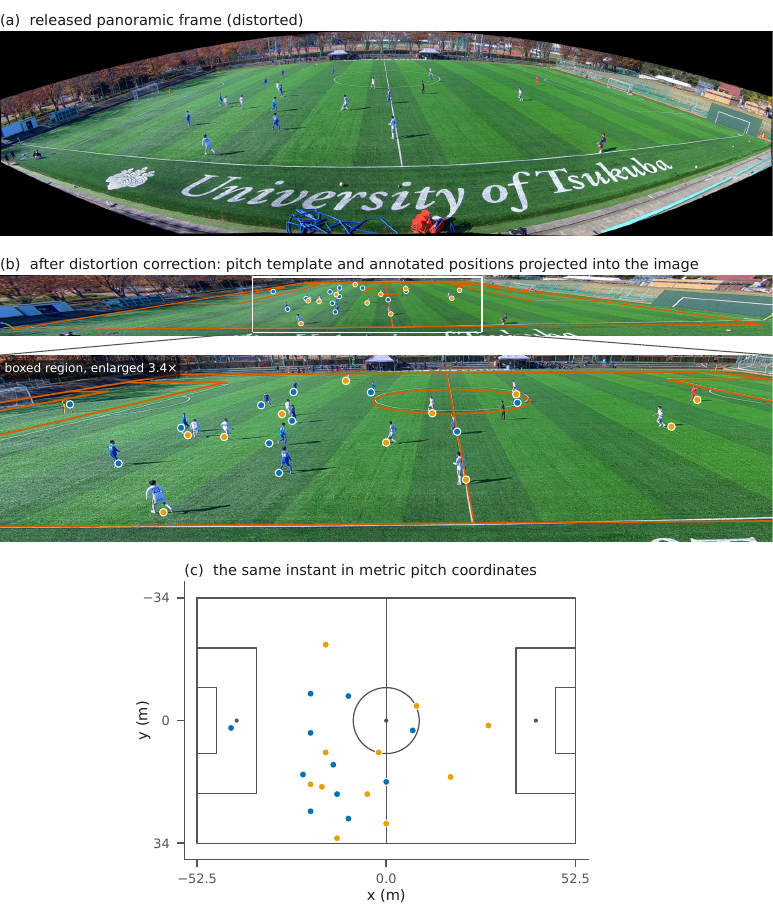}
  \caption{From released pixels to metric pitch coordinates, match M2,
  first half. (a) A frame of the released panoramic video: the stitched view
  is distorted and the pitch lines are curved. (b) The same frame after
  distortion correction, with the FIFA-standard pitch template and the
  frame's annotated player positions (coloured by team side) projected into
  the image through a homography fitted on the hand-annotated pitch
  keypoints, shown over the pitch region with the boxed region enlarged
  below. (c) The same instant in the metric coordinate system of the
  annotations: a $105 \times 68$\,m pitch with the origin at the centre
  circle, the touchline nearer the camera at the bottom.}
  \label{fig:calibration}
\end{figure}

Pitch coordinates are metric. The origin is the centre circle ($x = 0$,
$y = 0$); $x$ increases towards the right half of the pitch as seen from the
main camera, and $y$ increases towards the touchline nearer the camera. We
assume FIFA-standard dimensions of
$105\,\text{m} \times 68\,\text{m}$, so that goal lines are at $x = \pm 52.5$,
touchlines at $y = \pm 34.0$, and the halfway line at $x = 0$. Positions outside
the rectangle are legal, since balls and players can leave the field of play,
and are not clipped before metric computation, with one exception: in matches
M9 and M10 the recorded positions of the players as well as the ball
(Section~\ref{subsec:methods_bas_traj}) are clamped to the pitch rectangle at
source, so no position beyond the lines occurs in either.
Fig.~\ref{fig:calibration} illustrates the geometry: a released panoramic
frame, the same frame after distortion correction with the pitch template and
the frame's annotated positions projected into it, and the same instant in
the metric coordinate system of the annotations.

The release ships, for each match, $65$ hand-annotated correspondences
between pitch coordinates and the distortion-corrected panorama; they are
what Fig.~\ref{fig:calibration} and the pipeline's calibration stage
(Section~\ref{subsec:methods_gsr}) are fit from. A planar homography refit
to these correspondences reproduces them with a median residual of $2.9$ to
$6.9$\,px in nine of the ten matches; the exception is M1, a
training-split match whose stitch has a misaligned central seam, which no
single-view distortion model rectifies
(median $34.5$\,px). Expressed on the pitch through the inverse mapping, the
same correspondences show a median consistency of roughly $2$ to $3$\,m,
dominated by the far half of the pitch where a pixel subtends the most
metres; this is the scale against which the localisation component of
GS-HOTA, whose thresholds span $0.65$ to $5$\,m, should be read. Because the ground truth
and the pipeline's predictions reach pitch coordinates through different
calibrations, the provider's and a homography on these keypoints, part of
the localisation error is a mismatch between the two frames that we have not
quantified beyond this consistency check; on M2, ground-truth positions
projected through the keypoint homography land within a median of $14$\,px
of their annotated image positions. The released positions
themselves are quantised: every pitch coordinate lies on a fixed grid of one
hundredth of each pitch dimension, $1.05$\,m in $x$ and $0.68$\,m in $y$,
machine-exact across all twenty halves, so the grid sets the effective
positional resolution of the release. Beyond these consistency checks, the
provider's own positional accuracy cannot be audited from the release.

\subsection{Game state annotation}
\label{subsec:methods_gsr_annotation}

GSR annotations comprise 2D pitch locations for each player, persistent
player identities with their jersey numbers, team sides, and player roles. Positions and
identities were supplied by the BePro system synchronised to the video. We
inspected them visually against the
footage: the annotated positions were projected onto the panorama and onto a
bird's-eye view of the pitch (Figs.~\ref{fig:linkage}
and~\ref{fig:calibration}) and checked frame by frame in spot checks and
throughout the development of the baselines. No quantitative audit of
positional accuracy or identity consistency was performed; the consistency
checks of Section~\ref{subsec:methods_calibration} give the scale at which
the positions should be read.

GSR annotations are shipped in the SoccerNet GameState
format~\cite{soccergsr24}, one file per half, so that SoccerNet tooling
reads them directly. Each object annotation describes one player or
goalkeeper in one frame; the files contain no ball or referee annotations,
and each frame also carries normalised pitch-line points. An annotation
carries a per-half track identifier; a role (player or goalkeeper); a team
side, fixed per half from the attacking direction at kickoff, so one squad
may carry different sides in its two halves; the jersey number, present on
every record and constant along a track, so that it describes the player
rather than what is legible in the frame; the provider's integer player
identifier, present on every
record, which persists across the two halves of a match and, for squads that
appear in several fixtures, across matches, so it links a player's
appearances across the corpus without resolving them to a person; an
image-space box on the released panorama, placed from the pitch position with
a position-dependent size rather than drawn around each player, and distinct
from the human-annotated boxes of Section~\ref{subsec:methods_mot}; and the
pitch position, a single
point at the player's feet rather than a box. Jersey number and team side
are the identity the reconstruction benchmark scores; the event annotations
name their actor by the same player identifier.

\subsection{Event annotation}
\label{subsec:methods_bas_annotation}

Event data were obtained from the BePro system and converted to the twelve
classes used for ball action spotting. Classes are a many-to-one, hand-built mapping of the provider's event tags;
tags that carry no spottable ball action (possession spells, failed tackles
and duels, fouls, substitutions) are discarded. \texttt{Out}, which no BePro tag encodes, was
synthesised from the event stream itself: one \texttt{Out} is inserted before
each restart tag (throw-in, corner kick, free kick or successful goal kick),
placed one second after the preceding record when that record names an actor
and at its timestamp otherwise, and attributed to the last actor-bearing
action before the restart. It therefore marks play going dead rather than a
measured boundary crossing, and because free kicks are among the triggers,
roughly a fifth of \texttt{Out} events follow fouls rather than the ball
leaving the field.
The mapped timelines were then manually checked and corrected
against the synchronised video. Each event carries the acting team
and, when identifiable, the actor's player identifier, the same provider
identifier carried by the game state annotations, which is what allows
events to be cross-referenced with tracks.

BAS annotations are stored per match in the SoccerNet BAS
convention~\cite{cioppa2024soccernet}, with the half, a timestamp in
milliseconds from the first decoded frame of that half's video, and the
nearest video frame, which is the same frame index in the video and in the
game state annotation of that half. The
twelve
classes are \emph{Pass}, \emph{Drive},
\emph{Header}, \emph{High Pass}, \emph{Out}, \emph{Cross}, \emph{Throw In},
\emph{Shot}, \emph{Ball Player Block}, \emph{Player Successful Tackle},
\emph{Free Kick}, and \emph{Goal}. Dual actions at the same timestamp (for
example a shot that becomes a goal) are emitted as two separate annotations.

\subsection{Bounding-box subset curation}
\label{subsec:methods_mot}

Comprehensive bounding boxes are not part of the release. A curated subset
with bounding boxes and player identities is released instead, and was used
for the SoccerTrack Challenge at MMSports 2025: one four-minute panoramic
clip per match, $59{,}842$ clip frames in total, each annotated with the 22
players on the pitch, for $1{,}316{,}524$ boxes.

\subsection{Benchmark split}
\label{subsec:methods_split}

We adopt the SoccerTrack Challenge 2025 split, holding out M7 and M9
for test. How the eight training and validation matches are divided
between training and validation differs between the two
protocols the released tooling implements, and the difference matters
(see below). Every headline spotting figure in this paper comes from the
five-fold cross-match protocol of Section~\ref{subsec:methods_bas_traj},
under which the Challenge fold (fold 0) trains on six matches and validates
on M1 and M2. The released single-split protocol
instead reserves M2 and M10 for validation, a pair that mirrors the test
pair: each contains one of the two matches whose ball track is clamped to
the pitch rectangle. That pair enters the results through the
validation-sensitivity check below and through the earlier sweep that fixed
the network settings shared by all folds
(Section~\ref{subsec:methods_bas_traj}).

We deliberately do not split within matches. The two halves of a match share the two squads, both
squads' shape, the pitch, the camera and the weather, and every per-match data property, so a
within-match split would place all of those on both sides of the partition and conceal
exactly the variation a held-out match exposes.

The choice of \emph{validation} matches is equally consequential and less
obvious. With the test pair held at the Challenge matches, validating on
\{M2, M10\} rather than \{M1, M2\} lowers the with-ball mean average
precision at $\tau = 1$\,s from $0.617$ to about $0.40$ (three seeds,
$0.395$ to $0.408$), almost entirely on M9 ($0.53$ against $0.23$); M7 moves by
at most $0.03$. Both configurations select the same decoding threshold and
suppression radius, so the difference does not come from decoding. Changing
the validation pair also changes the training set: validating on \{M2, M10\}
trains on M1 in place of M10, the other match whose ball track is clamped,
and the loss falls on the clamped test match. We
therefore fix the validation rule as part
of the protocol, each fold validating on the following fold's test pair, and report the
cross-validated mean rather than any single configuration.

\subsection{Game state reconstruction baseline}
\label{subsec:methods_gsr}

\begin{figure}[!htbp]
  \centering
  \includegraphics[width=\linewidth]{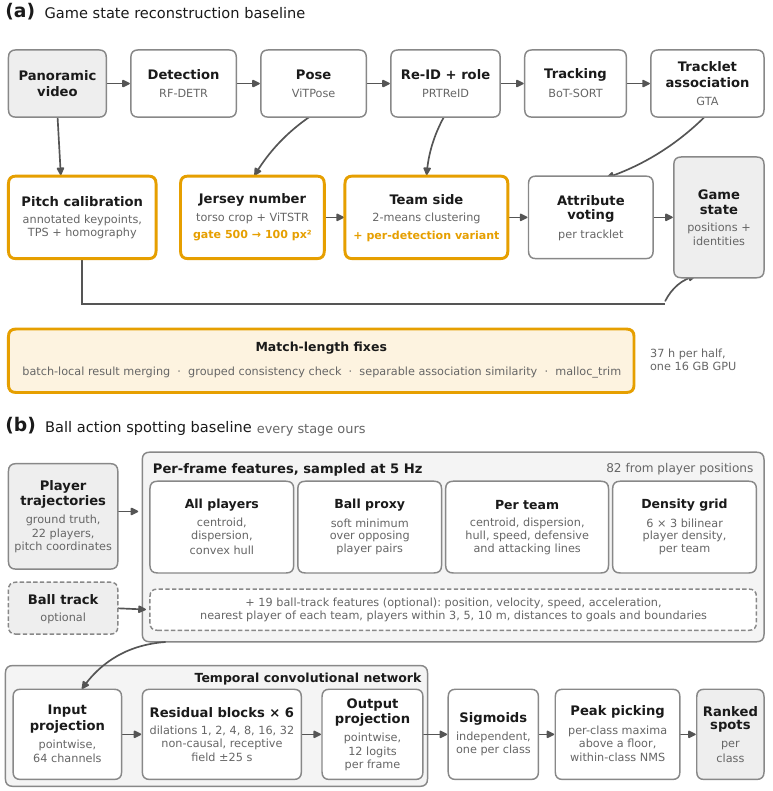}
  \caption{The two baselines. (a) Game state reconstruction. White boxes are
  published components; highlighted elements mark our contributions: the
  calibration stage fitted from the hand-annotated keypoints, the relaxed
  jersey-region gate, the per-detection team-assignment variant compared
  with the original on nested prefixes (Section~\ref{subsec:methods_gsr}),
  and the four fixes that let the pipeline run at match length
  (Section~\ref{subsec:methods_scale}). (b) Ball action spotting, every
  stage of which is ours (Section~\ref{subsec:methods_bas_traj}). Per-frame
  features computed from the supplied ground-truth player trajectories and
  sampled at $5\,\text{Hz}$, $82$ in four groups and optionally $19$ more
  from the ball track, pass through a non-causal temporal convolutional
  network: a pointwise projection to $64$ channels, six residual blocks with
  dilations $1$ to $32$ (receptive field $\pm 25\,\text{s}$) and a pointwise
  projection to twelve per-frame logits. Independent sigmoids score each
  class, and per-class peak picking above a floor with within-class
  non-maximum suppression yields ranked spots per class.}
  \label{fig:baselines}
\end{figure}

Our reference system is a game state reconstruction pipeline assembled in
the TrackLab framework~\cite{soccergsr24} from published components;
Fig.~\ref{fig:baselines}a diagrams it and marks our contributions. The
stages are RF-DETR~\cite{rfdetr25} person detection, ViTPose~\cite{vitpose22} pose estimation,
PRTReID~\cite{prtreid23} appearance features and role classification, BoT-SORT~\cite{botsort22}
tracking, pose-guided extraction of a torso region followed by ViTSTR~\cite{atienza2021vitstr}
jersey-number recognition, calibration by thin-plate
spline from the hand-annotated pitch keypoints to calibrated pixels and then a homography to
pitch coordinates (learned broadcast-view field registration~\cite{gutierrez2024nobells} is
trained on broadcast framing rather than fixed panoramas), global tracklet
association~\cite{sun2024gta}, and per-tracklet majority voting for role, team and jersey
number.
Beyond the published components, we contribute the calibration stage,
fitted per match from the hand-annotated correspondences of
Section~\ref{subsec:methods_calibration}; the relaxed jersey-region gate and
the per-detection team assignment described below; and the match-length
fixes of Section~\ref{subsec:methods_scale}. The
adaptation drew on released footage, including a test-split
half: the relaxed gate was first explored on prefixes of the first half of
M7 and the team-assignment comparison was made on the same prefixes.
Evaluation uses the official GS-HOTA metric~\cite{soccergsr24,luiten2021hota} in pitch
space with a $5\,\text{m}$ Gaussian tolerance. GS-HOTA partitions detections into classes by the
triple (role, team, jersey), so a detection carrying any incorrect attribute cannot match ground
truth at all rather than losing partial credit.

\paragraph{Jersey-region gating.}
The jersey stage crops a torso region from each detection using pose keypoints, then gates that
region before recognition on a minimum area and a minimum oriented-box aspect ratio. Those
thresholds are consequential at panoramic resolution. The median player box here is
$25 \times 55$\,px and the median proposed torso region $12 \times 11$\,px
($132\,\text{px}^2$), so the default gate of $500\,\text{px}^2$ at aspect ratio $0.6$ admits only
$6.9\%$ of detections. The recogniser is therefore starved of candidates rather than failing on
those it sees. We relax the area threshold to
$100\,\text{px}^2$, keeping the aspect ratio at $0.6$. A subsequent sweep on M2, one of
the two validation matches of the match-level split (Section~\ref{subsec:methods_split}),
ranks the same setting first; no sweep row is reported as a held-out result.

\paragraph{Team assignment.}
The default team assignment clusters tracklet-mean appearance embeddings with 2-means and labels the two sides
by mean pitch position. Because tracklets fragment as sequences lengthen, tracklet means become
unreliable, so we also implement clustering over per-detection embeddings with each tracklet
taking the majority label of its own detections. Goalkeepers, referees and non-person tracks are
excluded from clustering in both variants. Team accuracy is measured on
prediction and ground-truth pairs matched geometrically in pitch space by the Hungarian
assignment at a $2\,\text{m}$ threshold. Neither variant dominates on the nested prefixes of
Section~\ref{subsec:gsr_results}, so we report both there; every score of
Table~\ref{tab:gsr_results} uses the per-detection variant.

\subsection{Running the pipeline over a full half}
\label{subsec:methods_scale}

The pipeline's implementation is implicitly clip-scale. On a full half it does not finish, because of four scaling problems that
are not apparent at clip length. Each was individually blocking, the
fixes are documented with the released tooling so that the pipeline is runnable at the scale
the benchmark defines, and outputs were verified unchanged, exactly or to float32 rounding, in
every case.

\begin{itemize}
  \item \textbf{Quadratic result merging.} Per-batch result merging cost
        $O(N)$ per batch, so quadratic overall; restricting the merge to the
        batch's own rows and columns removes the superlinear term with
        score-identical output.
  \item \textbf{Quadratic consistency check.} A consistency check iterated
        the track-id column's values rather than its unique ids, giving
        $O(N^2)$; a single grouped pass has bit-identical output.
  \item \textbf{Association memory.} Global tracklet association
        materialised full $n_1 \times n_2$ similarity tensors on the GPU to
        compute a scalar mean, exceeding GPU memory over a 45-minute half;
        the mean pairwise cosine equals the dot product of mean unit
        embeddings, an exact separable identity.
  \item \textbf{Retained arena memory.} glibc retained freed per-thread
        arena memory across pipeline stages, so resident memory grew to tens
        of gigabytes while live data was under $1\,\text{GB}$ and the
        process was killed by the kernel; periodic \texttt{malloc\_trim}
        bounds the resident set.
\end{itemize}

With these changes a full half completes in approximately $37\,\text{h}$
on one $16\,\text{GB}$ GPU. Every half was run over its whole length and is
reported in Table~\ref{tab:gsr_results}.

Scoring long sequences needs one further robustness change. Calibration yields non-finite pitch
coordinates for a small and length-dependent fraction of detections near the
horizon, at most $0.78\%$ of object predictions over a full half. These
cannot be matched in pitch space; our
released scorer drops them and reports the count, where the official evaluator
aborts on the first such value.

\subsection{Trajectory-based ball action spotting}
\label{subsec:methods_bas_traj}

The ball action spotting baseline predicts, for each of the twelve event classes, a ranked
list of timestamps at which an event of that class occurs. We report temporal mean average
precision at a tight tolerance of $\tau = 1\,\text{s}$ and a loose tolerance of
$\tau = 5\,\text{s}$, following the SoccerNet ball action spotting
protocol~\cite{cioppa2023soccernet,cioppa2024soccernet}. We follow their implementation
exactly, which differs from the obvious reading in three ways that together change the score
by a factor of two to four: the tolerance $\tau$ is the \emph{total} width of the acceptance
window centred on the ground-truth event, so $\tau = 1\,\text{s}$ accepts a prediction within
$\pm 0.5\,\text{s}$ of it, not $\pm 1\,\text{s}$; assignment runs from each ground-truth event to the
highest-scoring unmatched prediction inside the window, not from predictions to the nearest
ground truth; and the precision-recall curve is sampled at $200$ fixed confidence thresholds
before $11$-point interpolation. Our evaluator is checked against a port of their source and
agrees to $10^{-9}$. Because the class distribution spans a factor of $300$ between the
most and least frequent class, we report both the unweighted mean over the twelve classes and
a mean weighted by class support, and we give the support alongside every per-class figure.
Average precision is computed over predictions pooled across the test matches rather than
averaged per match, so that a rare class is scored on all of its instances at once.

\paragraph{Benchmark definition.}
The benchmark comprises the $21{,}428$ events of Table~\ref{tab:bas_dataset},
of which $17{,}056$ fall in the eight training and validation matches and
$4{,}372$ in the two test matches.

\paragraph{Input representation.}
The model consumes the released ground-truth player positions on the pitch and no pixels;
predicted tracks are not used, so every figure in Section~\ref{subsec:bas_results} is an
offline result on the supplied trajectories. Positions are quantised to
$1.05\,\text{m}$ in $x$ and $0.68\,\text{m}$ in $y$, so a single-frame finite difference
carries $26\,\text{m}\,\text{s}^{-1}$ of quantisation noise; velocities are therefore computed
at two scales, $0.48\,\text{s}$ and $1.52\,\text{s}$, and both are supplied to the model. From
each frame we derive $82$ features in four groups: the centroid, dispersion and convex hull of
all $22$ players; a soft minimum over opposing player pairs, which serves as a proxy for the
location of the ball, since player positions constrain the ball strongly enough that its
trajectory can be inferred from them~\cite{kim2023ball}; per-team centroid, dispersion, hull, speed and defensive and attacking
lines; and a $6 \times 3$ bilinearly-binned player-density grid per team, which encodes
configuration without depending on any player ordering. Features are sampled at $5\,\text{Hz}$,
comfortably finer than the tighter evaluation tolerance.

We report two input conditions. The \emph{trajectory} condition uses the $82$ features above
and is reproducible from the released game-state annotations alone. The \emph{trajectory and
ball} condition adds $19$ features derived from the ball track, which is
released with the dataset as an auxiliary annotation. The ball track is a
reconstruction rather than a per-frame measurement: the ball moves in
straight lines at constant speed between anchors, with no deceleration, and
most anchors are annotated events, so most of its changes of velocity fall
within about $0.1$\,s of an annotated event instant, so this condition has access to information tied to the event
annotation. In matches M9 and M10 the recorded position is also clamped to
the pitch rectangle, as marked in Table~\ref{tab:bas_dataset}, and
Table~\ref{tab:bas_cv} gives the per-match consequence on the Challenge
fold. The features are: position,
velocity, speed, acceleration, distance to the nearest player of each team, the number of
players within $3$, $5$ and $10\,\text{m}$, and distances to each goal and to the boundary
lines. The per-frame ball-status flag is deliberately excluded: its out-of-play
value is close to a direct label for the \emph{Out} class on eight matches, and it is
constant on the remaining two.

\paragraph{Model, training and decoding.}
A temporal convolutional network~\cite{lea2017temporal} maps the feature
sequence to twelve per-frame logits (Fig.~\ref{fig:baselines}b). Its layout
follows the single-stage network of MS-TCN~\cite{farha2019mstcn}, with
residual blocks of two dilated convolutions~\cite{bai2018empirical}: a
pointwise input projection to $64$ channels, six residual blocks with dilations
$1$ to $32$ giving a receptive field of $\pm 25\,\text{s}$, and a pointwise output projection.
The network is non-causal, seeing $25\,\text{s}$ on either side of each
prediction, which is standard for offline action spotting and does not
describe a real-time system.
The twelve outputs are independent sigmoids rather than a softmax over classes, as in event
heads that allow simultaneous events~\cite{voeikov2020ttnet}, because
simultaneous events are part of the annotation convention. Targets are Gaussian of standard
deviation $300\,\text{ms}$ centred on each event, and the loss is binary cross-entropy with
per-class positive weighting capped at $50$. Training uses random $102\,\text{s}$ crops with
reflection augmentation in both pitch axes, which is label-preserving because no class is
handed. Models are selected on validation mean average precision, computed
with a looser $\pm\tau$ matching window than the reported metric, rather than
on validation loss, which the positive weighting renders non-monotone in the
reported metric. At inference,
per-class local maxima above a floor are thinned by within-class non-maximum suppression, and
each surviving peak is emitted with its height as its confidence. The floor,
the suppression radius and the stopping epoch within a fixed $24$-epoch
schedule are chosen on the validation matches of the fold being evaluated,
never on its test matches. The network width, dropout, learning rate and
schedule were fixed once, for all folds and both inputs, by an earlier sweep
of the trajectory model validated on M2 and M10, so for the two folds that
test on those matches one test match informed these shared settings.

\paragraph{Split.}
We evaluate with five-fold cross-match cross-validation. Each fold holds out two matches for
test and two for validation, every match appears in the test set exactly once, and fold~0 is
the SoccerTrack Challenge 2025 pair, so that fold~0 coincides with the held-out matches of
the Challenge and future work on that split is directly comparable with these figures.
Validation for each fold is the following fold's test pair, so fold~0 validates on M1 and
M2; the rule is arbitrary but fixed, and no fold's decoding or stopping
epoch is tuned on its own test matches. One
consequence is that fold~4 validates on the Challenge pair, so the cross-validated mean is not
fully independent of the Challenge fold, as in any $k$-fold design.

We do not split within matches, for the reasons given in
Section~\ref{subsec:methods_split}; for this release the shared per-match
properties are especially consequential, including the clamping of ball and
player positions in M9 and M10 described above. A fully team-disjoint
split is not achievable with these ten fixtures: one squad appears in four
matches and another in three, so some folds share a team between training
and test.

\paragraph{Chance baseline.}
Because two classes account for $82\%$ of events, and over the $932$ minutes an event
occurs on average every $2.6\,\text{s}$ and a \texttt{Pass} every $6\,\text{s}$
(Table~\ref{tab:bas_dataset}), evenly spaced guesses achieve a non-trivial score. We
therefore report a uniform-cadence chance baseline that emits spots at each
class's mean rate, estimated once on the eight non-test matches of the
Challenge split and used in every fold, and knows nothing else about the
match; estimating the rates on each fold's own training matches instead
changes the chance row by at most $0.002$.

\subsection{Evaluation metrics}
\label{subsec:methods_metrics}

For GSR we use the official SoccerNet GSR metric~\cite{soccergsr24}, with the
non-finite-coordinate handling described in Section~\ref{subsec:methods_scale}, and
report Game-State Higher Order Tracking Accuracy (GS-HOTA)~\cite{luiten2021hota}
with its standard decomposition into DetA, AssA and LocA. Scores are reported
per half at both lengths in Table~\ref{tab:gsr_results}, whose means are
over all twenty halves, with the mean over the four halves of the released
test split alongside. Supplementary
Table~\ref{tab:gsr_attrs_off} lists the DetA, AssA and LocA components of
the attributes-off scores at both lengths, so that every attributes-off
component quoted in Section~\ref{subsec:gsr_results} has a display item.
DetA and AssA are averaged over the metric's localisation thresholds, which
under its $5$\,m Gaussian kernel correspond to pitch distances from $0.65$
to $5$\,m, so DetA is not an accuracy at a single $5$\,m gate.

\paragraph{Evaluation windows.}
To measure how the score of one run depends on the length scored, the
whole-half predictions of every half are rescored against the first 30 seconds, 1, 2, 5, 10 and 20 minutes and
against the whole half; predictions and annotations outside the window are removed before
scoring, and non-finite coordinates are handled as above. The predictions
are those of Table~\ref{tab:gsr_results}, so the whole-half window
reproduces that table (Supplementary Table~\ref{tab:gsr_windows}). This is
distinct from the nested-prefix runs of Section~\ref{subsec:gsr_results},
which rerun the pipeline on each prefix of one half.

For BAS, a prediction at time $\hat{t}$ with class $c$ counts as a true positive
against a ground-truth event at time $t^{\star}$ with the same class if it
falls within a window of total width $\tau$ centred on $t^{\star}$, that is
$|\hat{t} - t^{\star}| \leq \tau/2$ (Section~\ref{subsec:methods_bas_traj}). We report a tight $\tau = 1\,\text{s}$
(maximum offset $0.5\,\text{s}$) and a loose $\tau = 5\,\text{s}$ (maximum
offset $2.5\,\text{s}$). Each ground-truth event matches at most one
prediction; remaining predictions count as false positives and remaining events
as false negatives. Dual events at the same timestamp are treated as two
independent targets. The reference evaluator implements the 11-point
interpolated AP computation.

The released evaluation tooling computes every metric reported here from
prediction files.

\subsection{Ethics and consent}
\label{subsec:methods_ethics}

All recordings were carried out under informed written consent agreements
approved by the ethical committee at the University of Tsukuba, and the consent
covers public release and redistribution of the recordings. The release is
pseudonymised: personal names are replaced by jersey
numbers, which are provided for re-identification and tracking purposes, and
the players' real-world identities are not included; because the video and the jersey numbers
are released together, individuals may remain identifiable to those who know
the teams.

\backmatter

\section*{Declarations}

\bmhead{Funding}
This work was financially supported by JST SPRING Grant Number JPMJSP2125 and JSPS KAKENHI Grant Number 23H03282.

\bmhead{Competing interests}
The authors declare no competing interests.

\bmhead{Ethics approval and consent to participate}
All recordings were carried out under informed written consent agreements
approved by the ethical committee at the University of Tsukuba (see
Section~\ref{subsec:methods_ethics}). The public release is pseudonymised:
personal names are replaced by jersey numbers, and the players' real-world
identities are not included. Because video and jersey numbers are released together, the data are
pseudonymised rather than anonymised. The written consent covers public release
and redistribution of the recordings.

\bmhead{Data availability}
The full package of videos, game state and ball action annotations, the
ball track used by the with-ball spotting condition, the
hand-annotated calibration keypoints and the curated bounding-box subset
(Section~\ref{subsec:methods_mot}) is hosted on Hugging Face at
\url{https://huggingface.co/datasets/atomscott/soccertrack-v2}, available
after accepting the dataset terms; the code and documentation are at
\url{https://github.com/AtomScott/SoccerTrack-v2}. The
release names matches by the provider's identifiers; its dataset card maps them
to the labels M1 to M10 used here. All results in this paper use release
v1.2 of the annotations, whose changes from v1.0 are listed in the dataset.
The data are
released under the CC BY 4.0 licence, which permits academic and commercial use
with attribution. Researchers are asked to cite this paper when publishing
results.

\bmhead{Code availability}
The dataset loader, the evaluation command-line tools for both metrics, and the
baseline starter kits are available at
\url{https://github.com/AtomScott/SoccerTrack-v2} under the MIT licence. The
evaluation tools compute every metric reported here from prediction files, and
the modifications to the reconstruction pipeline
(Sections~\ref{subsec:methods_gsr} and~\ref{subsec:methods_scale}) are
documented in the same repository.

\bmhead{Author contributions}
A.S.: conceptualization, methodology, software, validation, formal analysis,
investigation, data curation, visualization, writing (original draft) and
project administration. I.U., K.K. and Y.K.: data curation. K.F.:
conceptualization, supervision, funding acquisition and writing (review and
editing). All authors read and approved the final manuscript.

\begingroup\raggedright
\bibliography{refs}
\endgroup

\clearpage
\section*{Supplementary Information}
\label{sec:supplementary}
\setcounter{table}{0}
\renewcommand{\tablename}{Supplementary Table}

Supplementary Tables~\ref{tab:gsr_attrs_off} and~\ref{tab:gsr_windows} carry
the run-level detail behind Sections~\ref{subsec:gsr_results}
and~\ref{subsec:methods_metrics}: the DetA, AssA and LocA components of every
attributes-off score of Table~\ref{tab:gsr_results}, and the score of every
whole-half run over growing evaluation windows.

\begin{table}[!htbp]
  \centering
  \caption{Components of the attributes-off scores of
  Table~\ref{tab:gsr_results}: the same predictions rescored with
  attribute matching disabled, so that DetA, AssA and LocA measure
  geometry and association alone; \emph{Tracklets} is the number of
  predicted identities and \emph{Identities} the number of annotated
  identities in the sequence, substitutes included. The upper block is
  the opening 30 seconds of every half; the lower block is the full
  half of the same halves. Configuration is as in that table; the
  released test split is marked $\dagger$. Means are per column within
  each block, computed from unrounded scores.}
  \label{tab:gsr_attrs_off}
  \footnotesize
  \begin{tabular*}{\textwidth}{@{\extracolsep{\fill}}lrrrrr@{}}
    \toprule
    Half & DetA & AssA & LocA & Tracklets & Identities \\
    \midrule
    \multicolumn{6}{@{}l}{\emph{First 30\,s, all twenty halves}} \\
    M1, 1st & 21.60 & 35.39 & 73.05 & 24 & 22 \\
    M1, 2nd & 25.33 & 39.19 & 73.59 & 27 & 22 \\
    M2, 1st & 60.06 & 73.36 & 85.98 & 26 & 22 \\
    M2, 2nd & 41.67 & 58.57 & 82.69 & 33 & 22 \\
    M3, 1st & 57.79 & 66.38 & 85.65 & 25 & 22 \\
    M3, 2nd & 55.21 & 74.29 & 85.31 & 37 & 22 \\
    M4, 1st & 42.75 & 56.00 & 82.22 & 29 & 22 \\
    M4, 2nd & 49.24 & 71.02 & 83.66 & 25 & 22 \\
    M5, 1st & 66.28 & 78.77 & 88.34 & 29 & 22 \\
    M5, 2nd & 56.65 & 69.48 & 85.95 & 29 & 22 \\
    M6, 1st & 50.74 & 59.57 & 84.68 & 27 & 22 \\
    M6, 2nd & 50.30 & 58.85 & 84.67 & 28 & 22 \\
    M7, 1st$^\dagger$ & 73.68 & 81.63 & 89.56 & 26 & 22 \\
    M7, 2nd$^\dagger$ & 59.42 & 74.55 & 87.74 & 29 & 22 \\
    M8, 1st & 68.32 & 77.64 & 89.13 & 23 & 22 \\
    M8, 2nd & 57.54 & 64.63 & 85.09 & 23 & 22 \\
    M9, 1st$^\dagger$ & 26.37 & 33.67 & 77.84 & 28 & 22 \\
    M9, 2nd$^\dagger$ & 47.53 & 58.30 & 79.02 & 27 & 22 \\
    M10, 1st & 53.34 & 69.67 & 83.74 & 30 & 22 \\
    M10, 2nd & 33.63 & 58.19 & 81.17 & 29 & 22 \\
    Mean, all twenty halves & 49.87 & 62.96 & 83.45 & & \\
    \midrule
    \multicolumn{6}{@{}l}{\emph{Full half (45\,min), all twenty halves}} \\
    M1, 1st & 15.89 & 4.55 & 72.45 & 127 & 23 \\
    M1, 2nd & 17.01 & 3.27 & 72.63 & 137 & 33 \\
    M2, 1st & 45.98 & 10.84 & 82.28 & 146 & 23 \\
    M2, 2nd & 47.09 & 10.49 & 82.84 & 224 & 27 \\
    M3, 1st & 44.90 & 15.48 & 82.82 & 335 & 22 \\
    M3, 2nd & 46.01 & 12.40 & 83.20 & 422 & 29 \\
    M4, 1st & 46.26 & 13.24 & 83.01 & 141 & 23 \\
    M4, 2nd & 42.50 & 9.71 & 80.80 & 142 & 28 \\
    M5, 1st & 49.69 & 12.77 & 84.01 & 168 & 24 \\
    M5, 2nd & 46.05 & 13.39 & 83.64 & 351 & 30 \\
    M6, 1st & 49.43 & 10.80 & 84.15 & 123 & 22 \\
    M6, 2nd & 45.71 & 7.59 & 82.28 & 195 & 30 \\
    M7, 1st$^\dagger$ & 51.21 & 13.41 & 84.68 & 345 & 23 \\
    M7, 2nd$^\dagger$ & 50.82 & 12.52 & 84.84 & 293 & 23 \\
    M8, 1st & 49.15 & 13.76 & 84.30 & 133 & 23 \\
    M8, 2nd & 46.16 & 10.36 & 82.35 & 151 & 32 \\
    M9, 1st$^\dagger$ & 37.53 & 6.97 & 78.14 & 240 & 22 \\
    M9, 2nd$^\dagger$ & 41.42 & 7.77 & 77.61 & 273 & 28 \\
    M10, 1st & 46.43 & 18.63 & 83.30 & 240 & 22 \\
    M10, 2nd & 45.55 & 14.31 & 84.14 & 201 & 35 \\
    Mean, all twenty halves & 43.24 & 11.11 & 81.67 & & \\
    \bottomrule
  \end{tabular*}
\end{table}

\begin{table}[!htbp]
  \centering
  \caption{GS-HOTA of the whole-half predictions of
  Table~\ref{tab:gsr_results} scored over growing windows from kickoff:
  the first 30 seconds, 1, 2, 5, 10 and 20 minutes, and the whole half
  (Methods). The predictions are fixed and only the scored window
  changes, so the last column reproduces Table~\ref{tab:gsr_results}.
  The upper block is the official metric, the lower block the same
  windows with attribute matching disabled. Means are per column,
  computed from unrounded scores; the released test split is marked
  $\dagger$.}
  \label{tab:gsr_windows}
  \footnotesize
  \setlength{\tabcolsep}{4pt}
  \begin{tabular*}{\textwidth}{@{\extracolsep{\fill}}lrrrrrrr@{}}
    \toprule
    Half & 30\,s & 1\,min & 2\,min & 5\,min & 10\,min & 20\,min & Whole half \\
    \midrule
    \multicolumn{8}{@{}l}{\emph{Official GS-HOTA}} \\
    M1, 1st & 14.28 & 10.70 & 8.18 & 8.60 & 6.32 & 6.39 & 6.68 \\
    M1, 2nd & 9.90 & 7.08 & 5.48 & 5.77 & 3.92 & 4.14 & 3.60 \\
    M2, 1st & 29.15 & 21.02 & 17.24 & 17.44 & 16.87 & 17.01 & 14.28 \\
    M2, 2nd & 25.45 & 25.33 & 23.47 & 26.12 & 20.70 & 16.99 & 12.13 \\
    M3, 1st & 32.19 & 29.11 & 29.82 & 21.39 & 17.31 & 17.16 & 16.59 \\
    M3, 2nd & 25.36 & 26.86 & 23.33 & 17.97 & 16.95 & 13.62 & 12.21 \\
    M4, 1st & 34.05 & 31.15 & 28.14 & 22.10 & 16.61 & 13.56 & 12.36 \\
    M4, 2nd & 14.24 & 12.07 & 11.42 & 13.42 & 11.47 & 10.25 & 11.03 \\
    M5, 1st & 42.26 & 42.84 & 40.74 & 30.63 & 25.23 & 21.82 & 18.23 \\
    M5, 2nd & 29.34 & 30.77 & 30.72 & 23.19 & 18.63 & 21.35 & 16.54 \\
    M6, 1st & 11.90 & 13.32 & 17.28 & 14.89 & 11.90 & 12.87 & 13.36 \\
    M6, 2nd & 18.02 & 13.45 & 14.28 & 15.62 & 13.22 & 10.46 & 10.22 \\
    M7, 1st$^\dagger$ & 36.91 & 30.84 & 30.07 & 25.11 & 20.19 & 18.78 & 18.09 \\
    M7, 2nd$^\dagger$ & 24.90 & 24.64 & 21.87 & 26.05 & 20.12 & 16.33 & 12.89 \\
    M8, 1st & 24.68 & 19.21 & 16.22 & 15.53 & 11.84 & 12.61 & 12.73 \\
    M8, 2nd & 31.52 & 29.38 & 29.95 & 24.67 & 18.89 & 18.39 & 12.26 \\
    M9, 1st$^\dagger$ & 15.26 & 13.51 & 17.97 & 18.14 & 14.71 & 11.73 & 9.92 \\
    M9, 2nd$^\dagger$ & 5.84 & 6.60 & 6.34 & 4.81 & 3.73 & 3.64 & 2.94 \\
    M10, 1st & 27.91 & 28.26 & 27.51 & 27.82 & 27.37 & 19.64 & 15.27 \\
    M10, 2nd & 11.09 & 9.94 & 11.13 & 10.11 & 12.08 & 12.56 & 13.64 \\
    Mean, all twenty halves & 23.21 & 21.30 & 20.56 & 18.47 & 15.40 & 13.97 & 12.25 \\
    \midrule
    \multicolumn{8}{@{}l}{\emph{Attribute matching disabled}} \\
    M1, 1st & 26.30 & 18.58 & 13.54 & 12.53 & 9.75 & 8.53 & 8.46 \\
    M1, 2nd & 31.88 & 19.71 & 13.22 & 10.60 & 8.67 & 7.95 & 7.44 \\
    M2, 1st & 64.50 & 39.62 & 32.07 & 32.23 & 29.46 & 25.93 & 22.14 \\
    M2, 2nd & 49.16 & 52.22 & 48.28 & 42.51 & 35.45 & 28.05 & 22.08 \\
    M3, 1st & 61.77 & 53.21 & 53.68 & 39.81 & 32.27 & 28.90 & 26.09 \\
    M3, 2nd & 63.89 & 62.84 & 57.62 & 39.41 & 38.59 & 29.58 & 23.71 \\
    M4, 1st & 48.80 & 46.50 & 45.60 & 37.89 & 31.49 & 27.53 & 24.56 \\
    M4, 2nd & 58.32 & 51.66 & 38.49 & 31.05 & 26.37 & 20.30 & 20.05 \\
    M5, 1st & 71.79 & 71.66 & 60.85 & 45.30 & 37.16 & 32.67 & 25.10 \\
    M5, 2nd & 62.35 & 65.43 & 53.94 & 39.05 & 31.64 & 32.20 & 24.68 \\
    M6, 1st & 55.68 & 54.28 & 57.36 & 36.43 & 27.27 & 23.93 & 23.00 \\
    M6, 2nd & 53.17 & 39.57 & 38.29 & 30.78 & 24.25 & 21.93 & 18.49 \\
    M7, 1st$^\dagger$ & 75.81 & 60.53 & 54.81 & 37.18 & 30.86 & 29.08 & 26.04 \\
    M7, 2nd$^\dagger$ & 65.51 & 61.41 & 47.03 & 51.25 & 38.78 & 33.67 & 25.08 \\
    M8, 1st & 70.38 & 54.96 & 44.80 & 46.51 & 33.57 & 27.48 & 25.89 \\
    M8, 2nd & 61.06 & 55.20 & 54.60 & 42.95 & 35.98 & 31.46 & 21.65 \\
    M9, 1st$^\dagger$ & 29.54 & 23.25 & 23.87 & 24.52 & 21.14 & 18.13 & 16.15 \\
    M9, 2nd$^\dagger$ & 51.95 & 46.73 & 41.16 & 28.80 & 24.22 & 23.07 & 17.91 \\
    M10, 1st & 60.83 & 55.87 & 51.02 & 49.13 & 46.00 & 38.12 & 29.26 \\
    M10, 2nd & 43.87 & 49.15 & 51.53 & 43.05 & 34.67 & 28.88 & 25.40 \\
    Mean, all twenty halves & 55.33 & 49.12 & 44.09 & 36.05 & 29.88 & 25.87 & 21.66 \\
    \bottomrule
  \end{tabular*}
\end{table}

\end{document}